\documentclass[iicol,sn-mathphys-num]{sn-jnl}

\usepackage{graphicx}%
\usepackage{multirow}%
\usepackage{amsmath,amssymb,amsfonts}%
\usepackage{amsthm}%
\usepackage{mathrsfs}%
\usepackage[title]{appendix}%
\usepackage{xcolor}%
\usepackage{textcomp}%
\usepackage{manyfoot}%
\usepackage{booktabs}%
\usepackage{algorithm}%
\usepackage{algorithmicx}%
\usepackage{algpseudocode}%
\usepackage{listings}%

\usepackage{xcolor}
\usepackage{makecell} 

\graphicspath{
    {pics_exp/scene1/}
    {pics_exp/scene2/}
    {pics_exp/scene3/}
    {pics_exp/vel_pred/}
    {pics_exp/}
    {pics_methods/}
    }

\theoremstyle{thmstyleone}%
\theoremstyle{thmstyletwo}%

\theoremstyle{thmstylethree}%

\begin{document}

\title[Article Title]{Reliable and Real-Time Highway Trajectory Planning via Hybrid Learning-Optimization Frameworks}

\author[1]{\fnm{Yujia} \sur{Lu}}\email{23111256@bjtu.edu.cn}

\author*[2]{\fnm{Chong} \sur{Wei}}\email{weichong@hainanu.edu.cn}

\author[1]{\fnm{Lu} \sur{Ma}}\email{lma@bjtu.edu.cn}

\author[3]{\fnm{Lounis} \sur{Adouane}}\email{lounis.adouane@hds.utc.fr}

\affil[1]{\orgdiv{School of Traffic and Transportation}, \orgname{Beijing Jiaotong University}, \orgaddress{\city{Beijing}, \postcode{100044}, \country{China}}}

\affil*[2]{\orgdiv{School of Mechanical and Electrical Engineering}, \orgname{Hainan University}, \orgaddress{\city{Haikou}, \postcode{570228}, \state{Hainan}, \country{China}}}

\affil[3]{\orgdiv{Heudiasyc}, \orgname{Université de Technologie de Compiègne, CNRS}, \orgaddress{\city{Compiègne}, \postcode{60200}, \country{France}}}

\abstract{
Autonomous highway driving involves high-speed safety risks due to limited reaction time, where rare but dangerous events may lead to severe consequences. This places stringent requirements on trajectory planning in terms of both reliability and computational efficiency. This paper proposes a hybrid highway trajectory planning (H-HTP) framework that integrates learning-based adaptability with optimization-based formal safety guarantees. The key design principle is a deliberate division of labor: a learning module generates a traffic-adaptive velocity profile, while all safety-critical decisions including collision avoidance and kinematic feasibility are delegated to a Mixed-Integer Quadratic Program (MIQP). This design ensures that formal safety constraints are always enforced, regardless of the complexity of multi-vehicle interactions. A linearization strategy for the vehicle geometry substantially reduces the number of integer variables, enabling real-time optimization without sacrificing formal safety guarantees. Experiments on the HighD dataset demonstrate that H-HTP achieves a scenario success rate above 97\% with an average planning-cycle time of approximately 54 ms, reliably producing smooth, kinematically feasible, and collision-free trajectories in safety-critical highway scenarios.
}

\keywords{Autonomous Driving, Highway Trajectory Planning, Hybrid Learning–Optimization Framework, Mixed-Integer Quadratic Programming, Collision Avoidance}

\maketitle

\section{Introduction}\label{sec1:Introduction}

The trajectory planning module plays a central role in ensuring driving safety in autonomous driving systems, generating an optimal trajectory for the autonomous vehicle (AV) over a future time horizon based on structured environmental information from upstream perception and prediction modules. Safety in highway-oriented autonomous driving deserves particular attention: the high-speed operating conditions significantly amplify collision severity compared with urban traffic, and although structured road geometries reduce the frequency of chaotic interactions, the low-probability yet high-severity nature of highway accidents poses a critical planning challenge \cite{Zhang2025Integration,Cheng2025Safe,Zhang2023Enabling,Tang2022Highway}.

Trajectory planning approaches are broadly categorized as learning-based or optimization-based. Learning-based methods achieve strong generalization through data-driven modeling \cite{bojarski2016endtoend,kendall2018learning,hu2022planningoriented}, but struggle to provide reliable safety guarantees in rare yet critical situations. Optimization-based methods offer formal constraint modeling and safety guarantees \cite{Zhang2023Enabling,Li2023Real,Lim2021Hybrid}, but face computational challenges from the nonlinearity introduced by precise vehicle geometry and obstacle avoidance modeling \cite{Fan2024Exact,Schafer2023Computation}. This tension has motivated hybrid architectures that combine both paradigms \cite{Wang2022QPNet,Li2022Combining,Huang2024Differentiable, Feher2025RLPathPlanning,Li2024Trajectory}, yet constructing such frameworks for highway trajectory planning faces two major challenges.

The first challenge lies in \emph{data scarcity}. Extreme dangerous events on highways are inherently rare, resulting in a severe lack of real-world trajectory data for safety-critical scenarios \cite{Feng2024Curse}. This long-tail distribution undermines the training stability and coverage of data-driven approaches. Consequently, when learning modules are tasked with high-dimensional lateral decisions---such as lane-change initiation or full trajectory generation \cite{Huang2024Differentiable,Li2024Trajectory,Zhang2025Integration,Ni2024IntegratedRLDecision}---their training may fail to converge reliably in these critical cases. The second challenge concerns \emph{computational tractability}. To ensure reliable safety guarantees, optimization-based planners must explicitly model precise obstacle avoidance constraints. This requirement typically introduces high mathematical nonlinearity, as vehicle geometries and feasible regions are modeled as non-convex polygons \cite{Fan2024Exact,Schafer2023Computation}. Such formulations are difficult to solve within the stringent real-time requirements of highway driving, particularly when the number of integer variables in mixed-integer formulations scales unfavorably with the planning horizon \cite{Wang2024Convex}.

To address these challenges, we propose a hybrid highway trajectory planning (H-HTP) framework that introduces a principled division of labor between learning and optimization. Rather than tasking the learning module with high-dimensional lateral decision-making, we restrict it to longitudinal velocity planning, where human driving behavior exhibits stronger statistical regularity and is more amenable to data-driven modeling. All safety-critical decisions, including collision avoidance and kinematic feasibility, are delegated to a downstream optimization-based path planner. Within this planner, we introduce a linearization strategy that transforms the originally non-convex obstacle avoidance constraints into a compact mixed-integer quadratic program (MIQP) with a minimal number of integer variables, achieving real-time performance without sacrificing formal safety guarantees. Multi-agent traffic interactions are captured through a graph neural network built on a vectorized encoding scheme \cite{gao2020vectornet}, providing a generalizable velocity prior that guides the downstream planner across diverse highway scenarios.

The main contributions of this paper are as follows:
\begin{enumerate}
\item We propose H-HTP, a hybrid trajectory planning framework for highway driving that assigns longitudinal velocity planning to a learning module and lateral path planning to a constrained optimization module. This design ensures formal collision-avoidance guarantees while maintaining scalability to complex multi-vehicle interactions.
\item We introduce a linearization-based obstacle avoidance modeling strategy that reduces the non-convex vehicle geometry problem to an MIQP with a single integer variable per time step. Combined with a discretized vehicle body representation and a dynamic safety corridor, the formulation achieves real-time planning with high solution success rates.
\end{enumerate}
Experiments on real-world highway data from the HighD dataset \cite{krajewski2018highD} demonstrate that H-HTP reliably produces smooth, kinematically feasible, and collision-free trajectories in safety-critical scenarios, achieving a scenario success rate above 97\% with an average planning-cycle time of approximately 54~ms.\footnote{The complete source code is publicly available at \url{https://github.com/alasjia/H-HTP-Framework}.}

The remainder of this paper is organized as follows: Section~\ref{sec2:Related work} reviews related works. Section~\ref{sec3:Methods} presents the overall framework, with Sections~\ref{subsec3_1:traj_repre}--\ref{subsec3_3:vel_planning} detailing the trajectory representation, path planning, and velocity planning modules, respectively. Section~\ref{sec:experiments} reports experimental results, and Section~\ref{sec:conclusions} concludes the study.

\section{Related work}\label{sec2:Related work}

Trajectory planning for autonomous vehicles has been studied extensively from two complementary perspectives. The first concerns how to formally model collision avoidance, i.e., how to represent vehicle geometry and spatial constraints in a tractable yet reliable way. The second concerns the broader planning architecture, specifically how optimization-based and learning-based methods can be combined to balance safety guarantees with generalization. We review each perspective in turn and highlight the limitations that motivate the proposed H-HTP framework.

\subsection{Modeling of Collision Avoidance}\label{subsec2_1}

In trajectory planning, collision avoidance is commonly enforced through formally verified constraints that explicitly model the spatial relationships between vehicles. One line of research focuses on exact geometric computation techniques \cite{Fan2024Exact,Zhang2021Optimization}, which employ symbolic representations to compute intersections, distances, and spatial arrangements with mathematical exactness. These methods provide strong  rigor and completeness, and have demonstrated effectiveness in spatially constrained scenarios, such as parking maneuvers with limited mobility or navigation through narrow road segments.

To reduce modeling complexity, a large body of work introduces geometric approximations of vehicle shapes. Representative examples include elliptical representations \cite{Fan2024Efficient} and multi-circle models \cite{Liu2024Fast,Sun2022Successive}, which approximate vehicle geometry while retaining essential collision characteristics. However, constructing accurate collision avoidance constraints based on such geometric models often leads to high mathematical complexity \cite{Fan2024Exact,Schafer2023Computation}. To strike a practical balance between modeling fidelity and computational efficiency, several studies trade exactness for tractability by allowing bounded numerical approximations. Li et al. \cite{Li2023Real} reformulate polygon-based vehicle collision avoidance using a dual distance formulation, converting originally non-convex geometric constraints into linear constraints that can be efficiently solved within a convex MPC framework. Liu et al. \cite{Liu2024Fast} and Sun et al. \cite{Sun2022Successive} approximate non-convex collision avoidance constraints by iteratively constructing convex feasible sets around the current trajectory, achieving high planning success rates in practice. Han et al. \cite{Han2023RDA} propose a regularized dual alternating direction method (RDA), which solves a smooth bi-convex reformulation and enables parallel constraint computation for individual obstacles. Schafer et al. \cite{Schafer2023Computation} leverage reachability analysis to decompose non-convex free space into convex subsets and reformulate collision constraints through duality-based approaches.

However, in highway scenarios, relying solely on strictly formulated spatial collision-avoidance constraints is often insufficient. Highway traffic is characterized by rapid dynamics and complex behavioral interactions arising from numerous surrounding human-driven vehicles. Such properties can undermine safety guarantees derived purely from static spatial constraints. To address the uncertainty induced by highly dynamic traffic environments, several studies adopt probabilistic approaches to explicitly quantify uncertainty and support more conservative and robust decision-making. A popular line of recent work leverages artificial potential field (APF)–based methods to provide reference safety regions for the ego vehicle \cite{Yan2023Cooperative,Rousseas2022Trajectory,Huang2020Motion}. These approaches encode collision risk into potential functions, enabling reactive and computationally efficient safety-aware motion generation.

Beyond potential-field formulations, other studies explicitly quantify collision risk by incorporating perception and prediction uncertainty. For example, Alao et al. \cite{Alao2025Reliable} introduce stochastic predictive distance metrics that map uncertainty into probabilistic vehicle–agent distance risks, thereby transforming potential collision risks into computable safety indicators that can be incorporated into planning and control. In addition, some research focuses on modeling complex interactions between the ego vehicle and surrounding traffic participants. Representative efforts include game-theoretic frameworks \cite{Zhang2024Path,Chen2023Interaction} that explicitly model strategic interactions among agents, as well as approaches that emphasize accurate behavior prediction of surrounding vehicles \cite{Cheng2025Safe,Zhou2024Interaction} to mitigate uncertainty arising from complex interactive behaviors.

\subsection{Hybrid Trajectory Planning}\label{subsec2_2}

Trajectory planning approaches can be broadly categorized as optimization-based or learning-based. Optimization-based methods offer transparent constraint modeling and formal safety guarantees but often lack the flexibility to generalize across diverse traffic scenarios \cite{Zhang2023Enabling,Li2023Real,Lim2021Hybrid}. Learning-based methods, by contrast, demonstrate strong generalization through data-driven modeling but typically cannot provide reliable safety guarantees in rare or safety-critical situations \cite{bojarski2016endtoend,kendall2018learning,hu2022planningoriented}. This fundamental tension has motivated growing interest in hybrid architectures that combine the strengths of both paradigms.

Within hybrid frameworks, two interrelated design choices largely determine the system's safety properties and computational profile. The first concerns the \emph{division of labor} between the learning component and the optimization-based planner. Several works task the learning module with high-level lateral decisions such as lane-change initiation or target lane selection \cite{Li2022Combining,Zhang2025Integration,Li2024Trajectory}, while others predict full two-dimensional trajectories that are subsequently refined or verified \cite{Huang2024Differentiable,Wang2023DrivingStyleAwarePlanning}. A common architectural variant decouples path and speed planning: sequential approaches either generate a velocity profile along a precomputed path \cite{Wang2024Convex,Lim2021Hybrid} or plan a spatial path conditioned on a predefined speed reference \cite{Zhang2023Enabling,Yan2023Cooperative}; simultaneous approaches jointly optimize both dimensions at higher computational cost \cite{Zhang2024Path,Xiong2023Integrated,Lin2024Autonomous,Liu2022Dynamic}. However, when learning modules are responsible for lateral or full-trajectory decisions, the resulting outputs are difficult to verify for safety---a limitation that becomes critical under the data scarcity characterizing rare highway events. For instance, Cheng et al.\ \cite{Cheng2025Safe} enforce safety primarily through risk-based penalty terms rather than hard constraints, while the integrated planning and learning algorithm in \cite{Zhang2025Integration} does not enforce explicit trajectory-level collision avoidance.

The second design choice concerns the \emph{optimization formulation} used within the planning module. Nonlinear programming (NLP) can directly encode non-convex vehicle geometry and obstacle constraints \cite{Fan2024Exact}, but typically relies on local solvers without global optimality guarantees or predictable computation times. Sequential convex programming (SCP) \cite{Liu2024Fast,Sun2022Successive} iteratively constructs convex approximations, achieving practical success at the cost of multiple solver calls per cycle. Pure quadratic programming (QP) enables efficient single-pass computation and has been widely adopted \cite{Wang2022QPNet,Li2022Combining}, but maintaining convexity requires either simplified vehicle geometry---such as the implicit lane-based collision avoidance in QPNet \cite{Wang2022QPNet}---or relaxed safety constraints. Mixed-integer formulations address this trade-off by encoding discrete geometric or logical decisions through integer variables, while preserving a convex continuous relaxation amenable to mature branch-and-bound solvers \cite{Wang2024Convex,Liu2025TwoStageOptimizationPlanning}. Among these, MIQP combines the modeling flexibility of integer variables with quadratic smoothness objectives, making it particularly suitable for trajectory planning that requires both discrete geometric reasoning and continuous motion quality. However, the computational tractability of MIQP depends critically on the number of integer variables, which in prior work \cite{Wang2024Convex} scales with both the number of piecewise-linear segments and the planning horizon, potentially limiting real-time applicability.

In summary, the learning components in prior hybrid frameworks are typically tasked with high-dimensional lateral decision-making, which is prone to instability when training data for safety-critical scenarios is scarce. Many existing methods further rely on soft or indirect constraints, making it difficult to provide formal safety guarantees. Additionally, while MIQP offers a principled formulation for combining discrete geometric reasoning with continuous trajectory optimization, its real-time applicability has been constrained by excessive integer variables. The proposed H-HTP addresses these limitations on all three fronts: it restricts the learning module to the more statistically regular task of longitudinal velocity planning; it delegates all safety-critical decisions to a constrained optimization layer with formal collision-avoidance guarantees; and it introduces a linearization strategy that reduces the integer variable count to one per time step, enabling real-time MIQP execution without sacrificing modeling fidelity.

\section{Hybrid Highway Trajectory Planning Framework (H-HTP)}\label{sec3:Methods}

\begin{figure*}
    \centering
    \includegraphics[width=\textwidth]{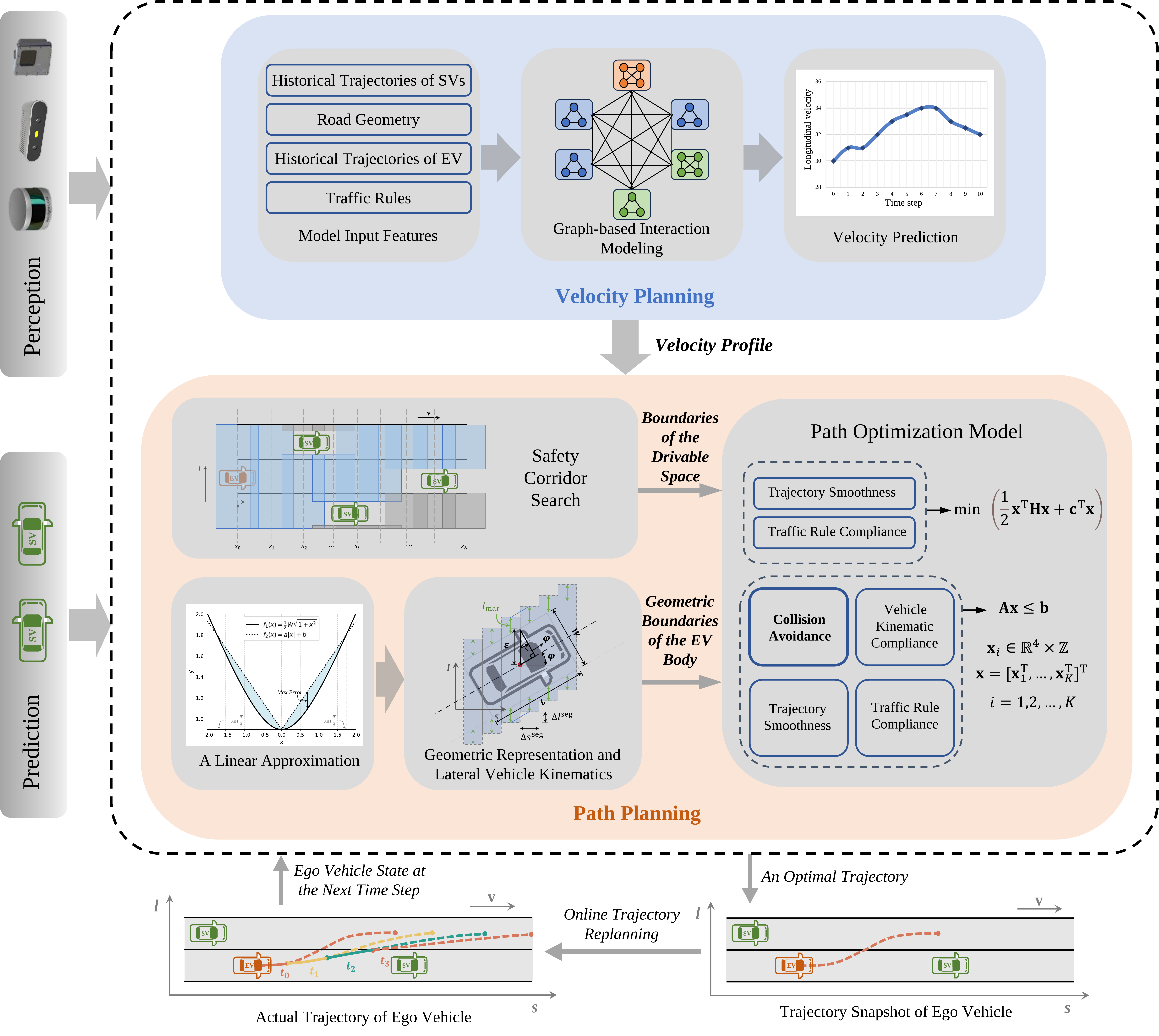}
    \caption{Overall framework of hybrid highway trajectory planning, comprising a learning-based velocity planner, an MIQP-based path planner with safety corridor construction and discretized vehicle geometry, and an online replanning module.}
    \label{fig:framework}
\end{figure*}

Fig.~\ref{fig:framework} illustrates the overall architecture of H-HTP. The central design principle of this framework is the deliberate placement of a constrained optimization-based planner as the downstream module responsible for all safety-critical decisions, including spatiotemporal collision avoidance and kinematic feasibility enforcement. This ensures that the final executed trajectory always satisfies formal safety guarantees, regardless of the upstream learning module's output. Upstream, a learning module generates a longitudinal velocity profile. The predicted velocity profile serves as a behavioral prior that informs, but does not override the downstream optimization. This asymmetric division of labor distinguishes H-HTP from prior hybrid frameworks, where learning components are often tasked with high-dimensional lateral decisions that are both data-hungry and difficult to verify for safety.

The remaining modules of the framework operate as follows. At each planning cycle, upstream perception and prediction modules provide the historical trajectories of the Ego Vehicle (EV) and Surrounding Vehicles (SVs), road geometry, traffic rules, and predicted SV motions. The learning-based velocity planner takes this contextual information as input and outputs a future longitudinal velocity profile for the EV, conditioned on the surrounding traffic context. This profile is passed to the downstream path planner as a longitudinal motion reference.

The path planning module, which constitutes the technical core of H-HTP, formulates the lateral trajectory generation as a constrained optimization problem. To efficiently handle dynamic interactions with surrounding vehicles, a heuristic safety corridor is first constructed to define the feasible drivable region over the planning horizon. Within this region, spatiotemporal non-overlapping constraints between the EV and SVs are explicitly imposed using a discretized geometric representation of the EV body. The resulting problem is solved as a mixed-integer quadratic program (MIQP), incorporating vehicle kinematic feasibility, motion smoothness, and traffic-rule compliance as additional constraints and objectives. The choice of MIQP as the optimization formulation is motivated by the problem structure that emerges from this design: the linearized vehicle geometry introduces a single binary sign variable per time step, while all remaining constraints are affine and the objective is purely quadratic---a structure that MIQP solvers exploit efficiently.

The remainder of this section details the technical components of H-HTP. Section~\ref{subsec3_1:traj_repre} introduces the trajectory representation and vehicle kinematic model, which form the mathematical foundation for subsequent formulations. Section~\ref{subsec3_2:path_planning} then presents the path planning module in full, covering the safety corridor construction, vehicle geometric representation and linearization, and the complete MIQP formulation—these three components are tightly coupled and together constitute the core technical contribution of this work. Finally, Section~\ref{subsec3_3:vel_planning} describes the learning-based velocity prediction module, focusing on the design choices that enable its seamless integration with the downstream optimizer.

\subsection{Trajectory Representation and Vehicle Kinematics}\label{subsec3_1:traj_repre}

\begin{figure}
    \centering
    \includegraphics[width=1\linewidth]{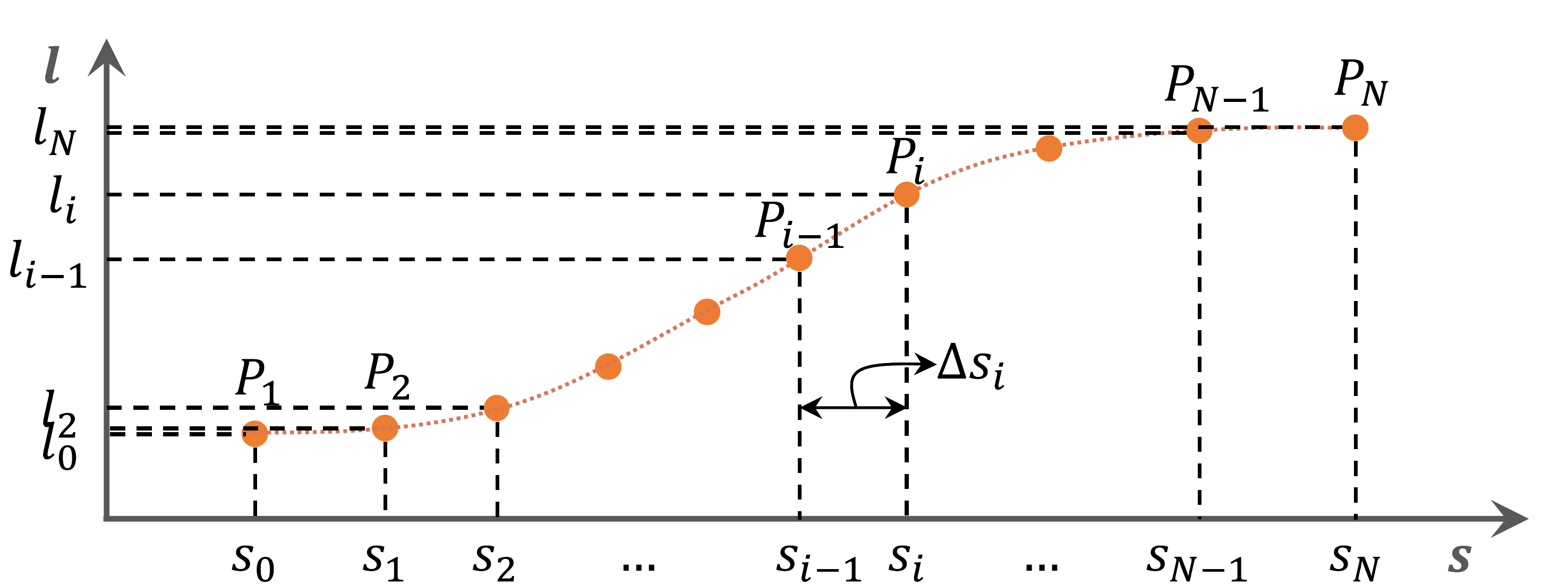}
    \caption{Discretization of Vehicle Trajectory}
    \label{fig:traj_discre}
\end{figure}

Typically, an s-l-t coordinate system is established to facilitate trajectory planning. In this system, the s-l coordinates define a Frenet frame \cite{werling2010frenet}, with the s-axis aligned with the road centerline and the l-axis orthogonal to it, while the t-axis corresponds to time. Discretizing the trajectory in this coordinate system is essential in autonomous driving, as it enables efficient computation and facilitates constraint handling and control. Accordingly, the target trajectory is represented as a sequence of discrete points in the s-l-t space. As illustrated in Fig.~\ref{fig:traj_discre}, the trajectory is defined as a sequence of discrete points:
\begin{equation}
    P = \{ P_i \}_{i=1}^{N}, \quad P_i = (s_i, l_i, t_i), \quad i = 1, 2, \ldots, N,
    \label{eq1:points}
\end{equation}
where $N$ denotes the total number of discretization steps and a constant time interval
\begin{equation}
    \Delta t = t_i - t_{i-1}
\label{eq2:delta_t}
\end{equation}
ensures uniform temporal resolution along the planning horizon.

This representation naturally decomposes the motion into longitudinal and lateral components, characterized by the sequences $\{s_i\}$ and $\{l_i\}$, respectively. Since Frenet coordinates can be transformed to and from the global Cartesian frame, the approach remains applicable to road segments with arbitrary curvature.

Although the planning framework operates on discrete trajectories, a continuous-time kinematic model is employed to describe the vehicle’s motion and to enforce dynamic constraints. The vehicle state at time~$t$ is defined as a tuple $(S(t), L(t), v(t), a(t), \varphi(t))$, where $S(t)$ and $L(t)$ denote the longitudinal and lateral coordinates of the vehicle's geometric center in the Frenet frame; 
$v(t)$ and $a(t)$ represent the instantaneous velocity and acceleration, respectively; and $\varphi(t)$ denotes the heading angle, i.e., the angle between the velocity vector and the $s$-axis. The system dynamics are governed by two additional quantities: the jerk $j(t) = \dot{a}(t)$ and the angular velocity $\omega(t) = \dot{\varphi}(t)$, yielding:

\begin{equation}
    \frac{d}{dt}
    \begin{bmatrix}
        S(t) \\
        L(t) \\
        v(t) \\
        a(t) \\
        \varphi(t)
    \end{bmatrix}
    =
    \begin{bmatrix}
        v(t) \cdot \cos \phi(t) \\
        v(t) \cdot \sin \phi(t) \\
        a(t) \\
        j(t) \\
        \omega(t)
    \end{bmatrix}
    \label{eq3:kinematics}
\end{equation}

\subsection{Path Planning via MIQP}\label{subsec3_2:path_planning}

Path planning is decomposed into three sequential stages. First, a heuristic safety corridor search identifies the obstacle-free drivable region along the EV's predicted longitudinal trajectory, converting the complex spatiotemporal collision-avoidance problem into a set of lateral boundary constraints. Second, a linearized geometric vehicle representation translates the EV's physical body shape into affine constraints suitable for convex optimization. Finally, these ingredients are assembled into a Mixed-Integer Quadratic Program that computes a dynamically feasible and smooth lateral path in real time.

\subsubsection{Safety Corridor Construction}\label{subsubsec:safe_corridor}

\begin{figure}
    \centering
    \includegraphics[width=1\linewidth]{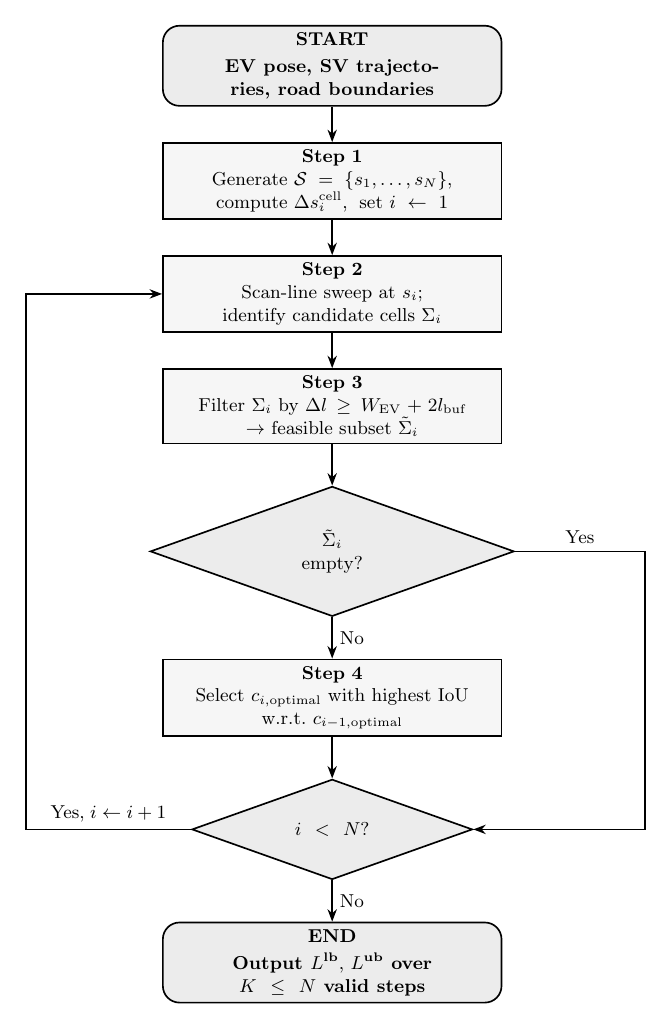}
    \caption{Flowchart of the Heuristic Safety Corridor Search Algorithm}
    \label{fig:corridor_flowchart}
\end{figure}

The safety corridor is constructed by identifying obstacle-free lateral intervals along the EV’s longitudinal reference positions and assembling them into a feasible drivable region over the planning horizon. The safety corridor is composed of discrete cells that represent optimal dynamic drivable regions for EV navigation. To efficiently identify these regions in real time, we employ a heuristic safety corridor search algorithm. This algorithm leverages available information, including the EV's current pose (position and orientation), its planned longitudinal motion, the current poses and predicted trajectories of SVs, road boundaries and lane markings. The search domain extends along the s-axis of the Frenet frame and terminates at the final predicted longitudinal position of the EV. Fig.~\ref{fig:safe_corridor} illustrates an example of the safety corridor search, where the blue cells denote the selected safety corridor, the shaded regions between adjacent cells indicate their lateral overlap ensuring corridor continuity, and the gray cells represent discarded candidates. The heuristic algorithm, whose workflow is summarized in Fig.~\ref{fig:corridor_flowchart}, works as follows:

\begin{figure}
    \centering
    \includegraphics[width=1\linewidth]{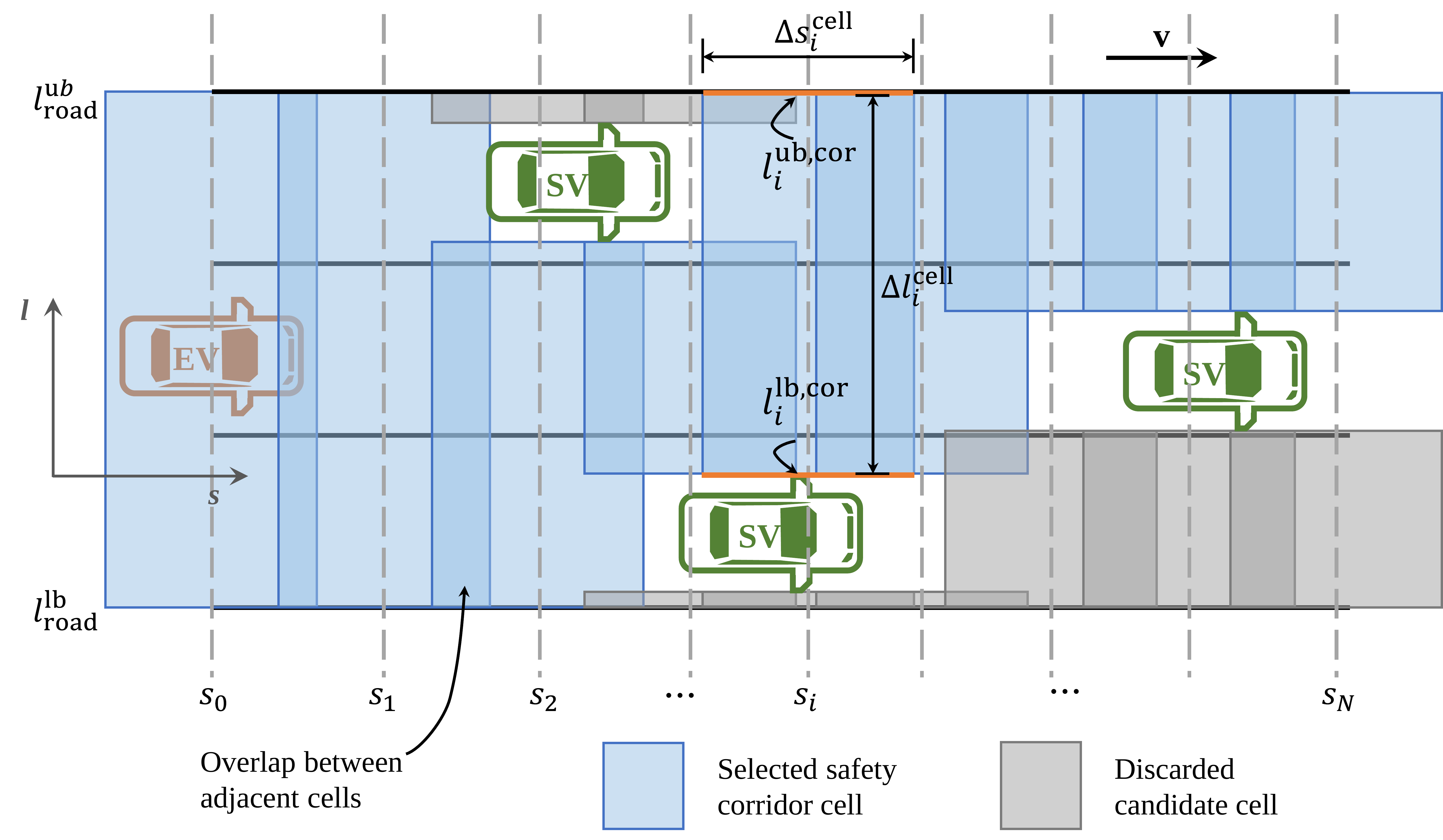}
    \caption{Illustration of safety corridor search. Blue cells: selected corridor; shaded regions: lateral overlap ensuring continuity; gray cells: discarded candidates.}
    \label{fig:safe_corridor}
\end{figure}

\medskip    
\textit{Step~1: Generation of Longitudinal Reference Positions}

A sequence of future longitudinal positions for the EV, denoted as
$\mathcal{S}=\{s_1,s_2,\dots,s_N\}$, is used as the center positions
of the safety cells distributed along the reference $s$-axis.
The corresponding longitudinal lengths of these cells,
$\{\Delta s_1^{\text{cell}},\Delta s_2^{\text{cell}},\dots,\Delta s_N^{\text{cell}}\}$,
are computed in parallel using \eqref{eq:delta_s_cell} defined later.

\medskip
\textit{Step~2: Obstacle-Free Cell Identification}

At each time step~$i$ corresponding to~$s_i$, a scan-line sweeps laterally along the $l$\nobreakdash-axis from $l_{\text{road}}^{\text{lb}}$ to $l_{\text{road}}^{\text{ub}}$, which are road boundaries.
Obstacles intersecting the scan-line divide it into multiple segments.
Each obstacle-free segment is stored as a rectangular cell $c_{i,d}\in\Sigma_i$, where
$\Sigma_i=\{c_{i,1},c_{i,2},\dots,c_{i,D_i}\}$ denotes the set of candidate cells at step~$i$.
The length of each cell $c_{i,d}$ is $\Delta s_i^{\text{cell}}$, and the width of each cell $c_{i,d}$ is
$\Delta l_{i,d}^{\text{cell}}=l_{i,d}^{\text{ub,cell}}-l_{i,d}^{\text{lb,cell}}$.

\medskip
\textit{Step~3: Feasible Cell Selection}

The candidate set~$\Sigma_i$ is filtered to obtain a feasible subset
$\tilde{\Sigma}_i\subseteq\Sigma_i$, containing only cells that are wide enough to accommodate the EV.
A cell $c_{i,d}\in\Sigma_i$ is retained if
$\Delta l_{i,d}^{\text{cell}}\ge W_{\text{EV}}+2l_{\text{buf}}$.
$W_{\text{EV}}$ and $l_{\text{buf}}$ represent the EV’s width and the lateral buffer distance required on each side of the vehicle, respectively.
If $\tilde{\Sigma}_i$ is empty, the process proceeds directly to \textit{Step~5}.

\medskip
\textit{Step~4: Optimal Cell Selection}

When multiple feasible cells exist in $\tilde{\Sigma}_i$, the 
selection criterion prioritizes \emph{spatial continuity} with 
respect to the previously selected cell $c_{i-1,\text{optimal}}$. 
This ensures the corridor evolves smoothly along the longitudinal 
axis rather than jumping discontinuously between separate lateral 
gaps, which would otherwise produce an infeasible or highly 
oscillatory planned path. Concretely, the cell with the highest 
Intersection-over-Union (IoU) overlap with $c_{i-1,\text{optimal}}$ 
is selected as a proximity measure, yielding the optimal cell 
$c_{i,\text{optimal}}$ spanning laterally over 
$[l_{i,\text{optimal}}^{\text{lb,cell}},\,
l_{i,\text{optimal}}^{\text{ub,cell}}]$.

\medskip
\textit{Step~5: Termination Condition}

If $i<N$, the process proceeds to \textit{Step~2} for time step $i+1$.
Otherwise, the search terminates.
The final output is a safety corridor defined by a sequence of lateral boundaries along the $l$\nobreakdash-axis:
$L^{\text{lb}}=\{l_1^{(\text{lb,cor})},l_2^{(\text{lb,cor})},\dots,l_K^{(\text{lb,cor})}\}$ and $L^{\text{ub}}=\{l_1^{(\text{ub,cor})},l_2^{(\text{ub,cor})},\dots,l_K^{(\text{ub,cor})}\}$, where each boundary satisfies
$l_i^{\text{lb,cor}}=l_{i,\text{optimal}}^{\text{lb,cell}}$ and $l_i^{\text{ub,cor}}=l_{i,\text{optimal}}^{\text{ub,cell}}$.

$K\le N$ is the number of valid time steps with achievable corridors. When $K < N$, the planning horizon is effectively shortened to cover only the $K$ steps for which a feasible corridor exists. The remaining $N - K$ steps are discarded for the current replanning cycle, and the shortened trajectory is passed downstream. This situation typically arises in heavily congested scenarios where no feasible lateral gap can be identified beyond a certain longitudinal distance.

\subsubsection{Vehicle Geometric Representation and Linearization}

\textbf{\textit{Linear Approximation}}

\begin{figure}
    \centering
    \includegraphics[width=1\linewidth]{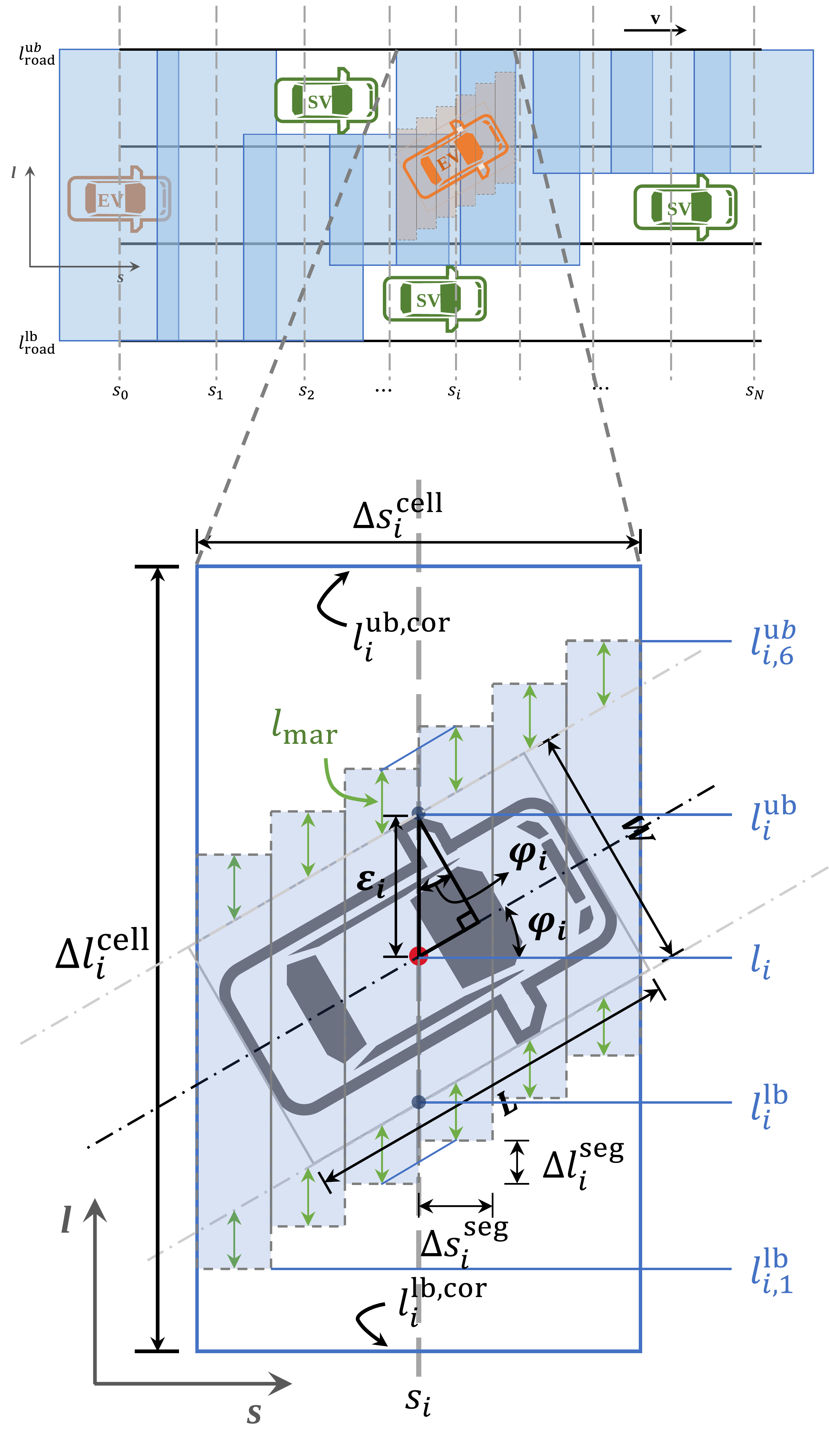}
    \caption{Geometric representation and lateral vehicle kinematics. The vehicle body is discretized into $\lambda$ segments, with lateral half-width offset $\varepsilon_i$ accounting for heading angle $\varphi_i$ and safety margin $l_{\mathrm{mar}}$ to corridor boundaries.}
    \label{fig:veh_geo}
\end{figure}

Fig.~\ref{fig:veh_geo} illustrates the vehicle geometric representation used for collision avoidance modeling. The upper panel shows the EV navigating among SVs within the safety corridor identified in Section~\ref{subsubsec:safe_corridor}, providing the spatial context in which the geometric constraints operate. The lower panel details the relationship between the EV’s body lateral boundaries $(l_i^{\text{ub}},l_i^{\text{lb}})$ and its lateral position~$l_i$:
\begin{equation}
\left\{
\begin{aligned}
l_i^{\text{ub}} &= l_i + \varepsilon_i \\[2pt]
l_i^{\text{lb}} &= l_i - \varepsilon_i
\end{aligned}
\right.,
\quad i = 1,2,\dots,K
\label{eq8:body_ublb}
\end{equation}
where $\varepsilon$ represents the lateral offset from the vehicle's geometric center to its boundary.  
The nonlinear dependence of $\varepsilon$ on the heading angle $\varphi$ arises from the fundamental kinematic relationship

\begin{equation}
\begin{split}
\tan\varphi &= \frac{dl}{ds}=l' \\
\Rightarrow\quad \varepsilon &= f_1(l') = \frac{1}{2}W_{\text{EV}}\sqrt{1+(l')^{2}},
\end{split}
\label{eq9:tan_epsilon}
\end{equation}
where $W_{\text{EV}}$ denotes the EV's width.  
This nonlinear formulation would render the path-planning optimization model nonlinear, significantly compromising computational efficiency.

\begin{figure}
    \centering
    \includegraphics[width=1\linewidth]{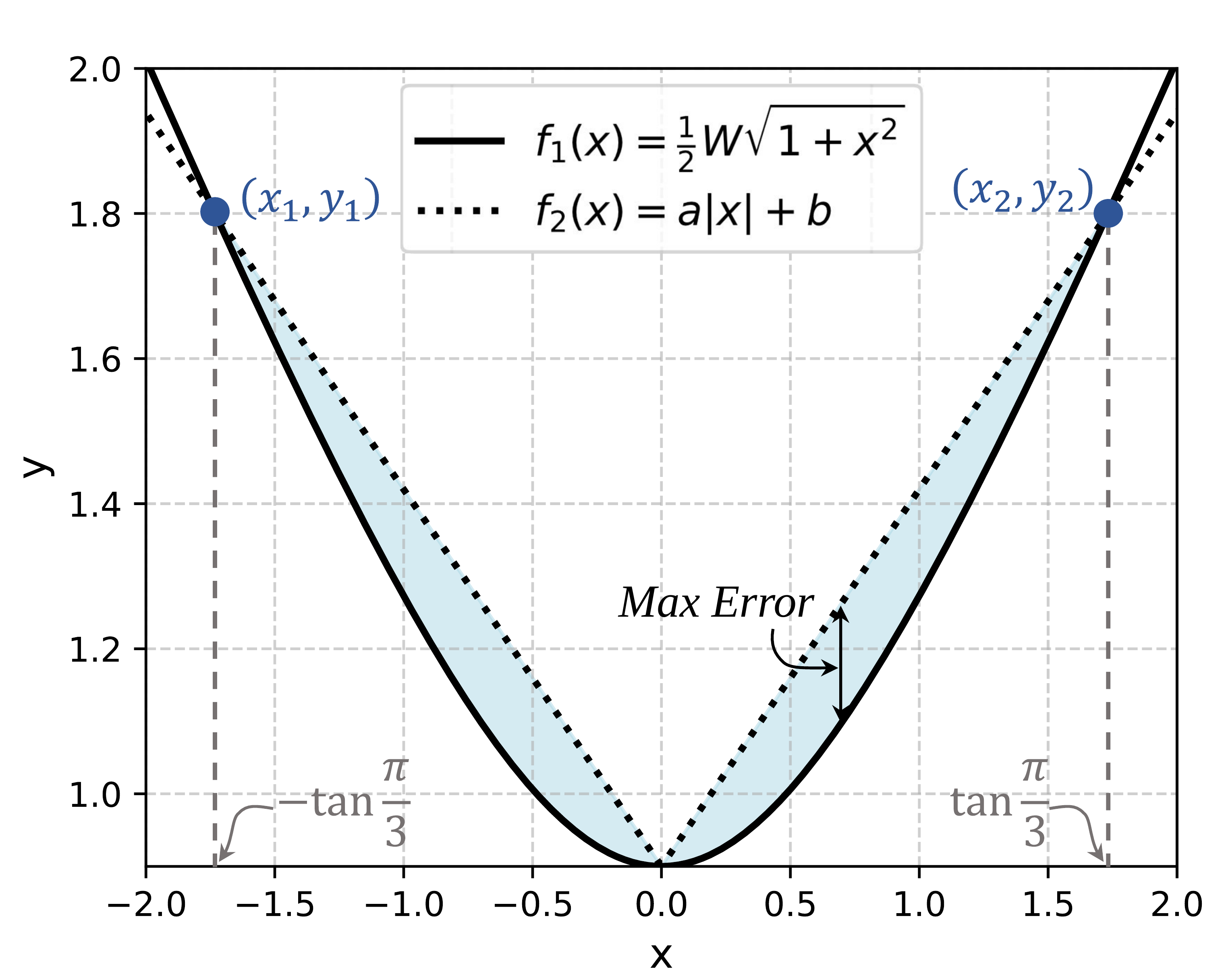}
    \caption{Piecewise-linear approximation $f_2(x)$ of the nonlinear half-width function $f_1(x)$, with a maximum error of 0.1652\,m over $|x| \leq \tan(\pi/3)$.}
    \label{fig:linear_appr}
\end{figure}

A linear approximation method is developed by leveraging the typically small steering angles in highway driving. Specifically, the maximum absolute EV heading angle is constrained to $|\varphi|\le \pi/3$, which yields $\tan\varphi\in[-\tan(\pi/3),\tan(\pi/3)]$. Fig.~\ref{fig:linear_appr} illustrates the piecewise-linear approximation of the absolute-value function constructed through endpoint fitting at $(x_1,y_1)$, $(x_2,y_2)$, and the origin $(0,0)$, where
\begin{equation}
\varepsilon = f_2(l') = a|l'| + b.
\label{eq10:epsilon_func}
\end{equation}

For a standard vehicle width $W_{\text{EV}} = \text{1.8~m}$, the derived parameters are $a \approx 0.5196$ and $b \approx 0.9$. Two properties make this approximation suitable for safety-critical planning. First, the inequality $f_2(l') \geq f_1(l')$ holds rigorously for all $l' \in [-\tan(\pi/3),\, \tan(\pi/3)]$, meaning the approximated geometry always encloses the actual vehicle contour. This ensures that any trajectory satisfying the linearized constraints is guaranteed to be collision-free with respect to the true vehicle shape. Second, the maximum approximation error is only 0.1652~m (Fig.~\ref{fig:linear_appr}), indicating that the additional conservatism introduced by the over-approximation is negligible in practice and does not unduly restrict the feasible solution space.

To implement the piecewise-linear approximation within the optimization framework, the formulation is convexified by introducing sign variables $k_i$, 
which encode the sign of $l'_i$ and are implemented as integer variables within the MIQP:

\begin{equation}
\varepsilon_i = a\,k_i\,l'_i + b,
\label{eq:epsilon_linear}
\end{equation}
with
\begin{equation}
k_i =
\begin{cases}
\phantom{-}1, & \text{if } l'_i \ge 0,\\[2pt]
-1, & \text{if } l'_i < 0.
\end{cases}
\label{eq:k_cases}
\end{equation}

\label{subsubsec:geometric_linear}

\textbf{\textit{Vehicle Geometric Representation}}

With $\varepsilon_i$ now expressed as the linear function of $l'_i$ in~\eqref{eq:epsilon_linear}, the upper and lower body boundaries $l_i^{\text{ub}} = l_i + \varepsilon_i$ and $l_i^{\text{lb}} = l_i - \varepsilon_i$ from~\eqref{eq8:body_ublb} become affine in the decision variables, enabling an efficient discrete geometric representation of the vehicle body. 
As illustrated in Fig.~\ref{fig:veh_geo}, the vehicle body is discretized into $\lambda=$~6 rectangular segments oriented along the longitudinal $s$-axis. The safety margin $l_{\mathrm{mar}}$ represents the minimum required lateral clearance between the EV body and the corridor boundaries. For each time step $i$, the length of each segment is determined by
\begin{align}
\Delta s_{i}^{\text{seg}} &= \frac{\Delta s_{i}^{\text{cell}}}{\lambda} \label{eq:delta_s_seg} \\
\Delta s_{i}^{\text{cell}} &= \omega_{\text{cell}}\,v_i \label{eq:delta_s_cell}
\end{align}
where $\Delta s_{i}^{\text{cell}}$ represents the longitudinal length of the $i$-th safety-corridor cell $c_{i,\text{optimal}}$, which fully encloses the vehicle body. This length scales proportionally with the EV's instantaneous velocity through the tunable parameter $\omega_{\text{cell}}$, which defines the temporal look-ahead duration used to size each cell. Specifically, $\Delta s_i^{\text{cell}}$ approximates the longitudinal distance the EV would travel in $\omega_{\text{cell}}$ seconds at speed $v_i$. A larger $\omega_{\text{cell}}$ yields longer cells that provide more conservative spatial margins around the vehicle body.

The lateral offset between adjacent segments is calculated as
\begin{equation}
\Delta l_{i}^{\text{seg}} = \Delta s_{i}^{\text{seg}}\tan\varphi_i 
                         = \Delta s_{i}^{\text{seg}}\,l_i'. \label{eq:delta_l_seg}
\end{equation}
This approach yields the following boundary constraints for the vehicle's geometric representation:
\begin{equation}
\left\{
\begin{array}{l}
l_{i,j}^{\text{ub}} = l_i^{\text{ub}} + \bigl(j-\tfrac{\lambda+1}{2}\bigr)\Delta s_{i}^{\text{seg}}l_i' \\[4pt]
l_{i,j}^{\text{lb}} = l_i^{\text{lb}} + \bigl(j-\tfrac{\lambda+1}{2}\bigr)\Delta s_{i}^{\text{seg}}l_i'
\end{array}
\right.,
\quad
\begin{array}{l}
i = 1,2,\dots,K;\\[4pt]
j = 1,2,\dots,\lambda.
\end{array}
\label{eq:body_ublb_contents}
\end{equation}

\subsubsection{MIQP Formulation for Path Optimization}
\label{subsubsec:path_opt_model}

Given the safety corridor boundaries and the vehicle 
geometric representation, the path planning 
problem is formulated as a constrained optimization problem. 
Specifically, given the current kinematic states of the EV and SVs, 
the road geometry, and the longitudinal velocity profile provided by 
the upstream learning module, the objective is to compute a 
dynamically feasible lateral path that:
\begin{enumerate}
    \item Enforces spatiotemporal collision-avoidance constraints 
    with respect to surrounding vehicles over the planning horizon,
    \item Satisfies the EV's kinematic limits and relevant traffic 
    rules, and
    \item Promotes smooth and continuous motion.
\end{enumerate}

\textbf{\textit{Decision Variables.}}
Following the MIQP-based path planning framework 
of~\cite{Wang2024Convex}, the optimization variable at each 
time step $i$ is defined as:
\begin{equation}
    \begin{aligned}
    &\mathbf{x}_i = \left( l_i, l_i', l_i'', l_i''', k_i \right), 
    \quad \mathbf{x}_i \in \mathbb{R}^4 \times \mathbb{Z}, \\
    &\quad i = 1, 2, \dots, K
    \end{aligned}
    \label{eq:var_x}
\end{equation}
where $l_i$ denotes the EV's lateral position at its geometric 
center; $l_i'$, $l_i''$, and $l_i'''$ are its first-, second-, and 
third-order derivatives with respect to $s$, capturing heading, 
curvature, and curvature rate respectively, with $l' = dl/ds = 
\tan\varphi$; and $k_i \in \{-1, +1\}$ is the sign variable introduced in 
Section~\ref{subsubsec:geometric_linear} to encode the sign of $l'_i$ 
in the piecewise-linear heading approximation. The full optimization 
variable is stacked as $\mathbf{x} = [\mathbf{x}_1^\top, \dots, 
\mathbf{x}_K^\top]^\top$, where $K \leq N$ is the number of valid 
time steps with achievable safety corridors.

\textbf{\textit{Full Problem Formulation.}}
The above objectives and constraints are jointly encoded within the 
following unified MIQP:
\begin{equation}
    \begin{aligned}
    \min \quad & \frac{1}{2} \mathbf{x}^\top \mathbf{H} \mathbf{x} 
    + \mathbf{c}^\top \mathbf{x} \\
    \mathrm{s.t.} \quad & \mathbf{A} \mathbf{x} \leq \mathbf{b} \\
    & \mathbf{x} = \begin{bmatrix} \mathbf{x}_1^\top, \dots, 
    \mathbf{x}_K^\top \end{bmatrix}^\top, \\
    & \mathbf{x}_i \in \mathbb{R}^4 \times \mathbb{Z}, 
    \quad i = 1, 2, \dots, K  \\
    \end{aligned}
    \label{eq:MIQP}
\end{equation}
Here $\mathbf{H} \in \mathbb{R}^{5K \times 5K}$ is a positive 
semi-definite matrix and $\mathbf{c} \in \mathbb{R}^{5K}$ is a 
linear cost vector, both assembled from the cost terms 
in~\eqref{eq:obj_cost}. The linear component $\mathbf{c}$ arises 
from the reference waypoint terms $l_{\mathrm{ref},i}$.
The constraint matrix $\mathbf{A}$ and vector $\mathbf{b}$ are 
assembled by stacking the following four groups of linear inequality 
constraints row-wise:
\begin{enumerate}
    \item \emph{Collision avoidance}: the vehicle geometry 
    constraints~\eqref{eq:coll_avoi_final_constraints}, enforced 
    for all $\lambda$ segments and $K$ time steps;
    \item \emph{Road boundaries}: lateral position 
    bounds~\eqref{eq:road_bound_cons};
    \item \emph{Kinematic continuity}: the Taylor expansion 
    equalities~\eqref{eq:kinematic_cons}, reformulated as 
    pairs of inequalities;
    \item \emph{Physical limits}: derivative 
    bounds~\eqref{eq:kinematic_cons2}.
\end{enumerate}
The mixed-integer structure arises from $k_i \in \{-1,+1\}$, 
the sign variable in each $\mathbf{x}_i \in \mathbb{R}^4 \times 
\mathbb{Z}$, with the remaining four components 
$(l_i,\,l_i',\,l_i'',\,l_i''')$ being continuous.
Note that the bilinear product $k_i\,l_i'$ appearing 
in~\eqref{eq:epsilon_linear}  is linearized using standard big-M techniques, replacing it with an auxiliary continuous variable and additional linear constraints to preserve the MIQP structure.

\textbf{\textit{Collision Avoidance Constraints.}}
Building directly on the discretized vehicle geometry 
from Section~\ref{subsubsec:geometric_linear}  and the safety corridor boundaries 
from Section~\ref{subsubsec:safe_corridor}, the spatiotemporal 
non-overlapping condition between the EV and SVs is enforced as:
\begin{equation}
    \left\{
    \begin{array}{l}
    l_{i,j}^{\mathrm{ub}} + l_{\mathrm{mar}} \leq 
    l_{i}^{\mathrm{ub,cor}} \\[2pt]
    l_{i,j}^{\mathrm{lb}} - l_{\mathrm{mar}} \geq 
    l_{i}^{\mathrm{lb,cor}}
    \end{array}
    \right.,
    \quad
    \begin{array}{l}
    i = 1,2,\dots,K;\\[2pt]
    j = 1,2,\dots,\lambda.
    \end{array}
    \label{eq:coll_avoi_final_constraints}
\end{equation}
where $l_{i,j}^{\mathrm{ub}}$ and $l_{i,j}^{\mathrm{lb}}$ are the 
upper and lower boundaries of the $j$-th geometric segment of the 
EV body at time step $i$, as defined in 
\eqref{eq:body_ublb_contents}; $l_{i}^{\mathrm{ub,cor}}$ and 
$l_{i}^{\mathrm{lb,cor}}$ are the safety corridor boundaries; and $l_{\mathrm{mar}}$ 
is the minimum required safety margin.

A key computational advantage of the proposed linearization is the significant reduction in combinatorial complexity. The approach in~\cite{Wang2024Convex} adopted a big-M formulation requiring $Q \times N$ integer variables, where $Q$ denotes the number of piecewise-linear segments. By contrast, the proposed $\pm 1$ sign variable encoding in Section~\ref{subsubsec:geometric_linear} requires only $N$ integer variables $k_i$ --- one per time step --- reducing the integer variable count by a factor of $Q$. Since MIQP solve time grows combinatorially with the number of integer variables, this reduction directly translates to faster real-time planning performance. 

By iterating over all 
$\lambda$ segments and $K$ time steps, these constraints enforce 
collision-free motion throughout the entire planning horizon while 
explicitly accounting for the EV's steering-aware body shape.

\textbf{\textit{Road Boundary Constraints.}}
The EV must remain within the road boundaries during highway 
driving. The road boundaries, defined by the lower boundary 
$l_{\mathrm{road}}^{\mathrm{lb}}$ and upper boundary 
$l_{\mathrm{road}}^{\mathrm{ub}}$, are obtained through perception 
modules or High-Definition maps. This constraint is formally expressed as:
\begin{equation}
    l_{\mathrm{road}}^{\mathrm{lb}} \leq l_i \leq 
    l_{\mathrm{road}}^{\mathrm{ub}}.
    \label{eq:road_bound_cons}
\end{equation}

\textbf{\textit{Vehicle Lateral Kinematic Constraints.}}

Motion continuity and physical feasibility are enforced via a 
third-order Taylor expansion about $s_{i-1}$ over the interval 
$\Delta s_i = s_i - s_{i-1}$. A third-order model is adopted 
because $l'''$, the rate of change of curvature with respect to 
$s$, directly governs the smoothness of steering transitions 
and prevents abrupt curvature jumps that would be 
dynamically infeasible. Initial conditions $(l_0, l_0', l_0'')$ 
are inherited from the previous replanning cycle to ensure 
continuity across successive planning horizons:
\begin{equation}
    \left\{
    \begin{aligned}
    l_i'' &= l_{i-1}'' + l_i''' \Delta s_i \\[2pt]
    l_i'  &= l_{i-1}' + l_i'' \Delta s_i 
             + \tfrac{1}{2} l_i''' (\Delta s_i)^2 \\[2pt]
    l_i   &= l_{i-1} + l_i' \Delta s_i 
             + \tfrac{1}{2} l_i'' (\Delta s_i)^2 
             + \tfrac{1}{6} l_i''' (\Delta s_i)^3
    \end{aligned}
    \right.
    \label{eq:kinematic_cons}
\end{equation}
The vehicle's physical limits further impose bounds on the state derivatives. The bound on $l'$ follows directly from the maximum heading angle constraint $|\varphi| \leq \pi/3$ established in Section~\ref{subsubsec:geometric_linear}, yielding $|l'| = |\tan\varphi| \leq \tan(\pi/3)$. The bound on $l''$ reflects the maximum lateral curvature achievable at the vehicle's minimum turning radius, while the bound on $l'''$ limits the rate of curvature change to ensure smooth and physically realizable steering transitions:
\begin{equation}
    \left\{
    \begin{aligned}
    l_{\min}' &\leq l_i' \leq l_{\max}' \\[2pt]
    l_{\min}'' &\leq l_i'' \leq l_{\max}'' \\[2pt]
    l_{\min}''' &\leq l_i''' \leq l_{\max}'''
    \end{aligned}
    \right.
    \label{eq:kinematic_cons2}
\end{equation}

\textbf{\textit{Objective Function.}}
The cost function balances lane centering, trajectory smoothness, 
and discretization regularity:
\begin{equation}
    \begin{aligned}
    \min\;
    &\omega_{1}\sum_{i=1}^{N}(l_{i}-l_{\mathrm{ref},i})^{2}
    +\omega_{2}\sum_{i=1}^{N}(l_{i}')^{2}\\
    &+\omega_{3}\sum_{i=1}^{N}(l_{i}'')^{2}
    +\omega_{4}\sum_{i=1}^{N}(l_{i}''')^{2}\\
    &+\omega_{5}\sum_{i=2}^{N}(l_{i}-l_{i-1})^{2}
    +\omega_{6}\sum_{i=2}^{N}(l_{i}'-l_{i-1}')^{2}
    \end{aligned}
    \label{eq:obj_cost}
\end{equation}
The term weighted by $\omega_1$ minimizes deviation from reference 
waypoints $l_{\mathrm{ref},i}$ to maintain optimal lane positioning, 
where $l_{\mathrm{ref},i}$ follows the lane centerline when markings 
are available and defaults to the corridor center otherwise. In 
multi-lane scenarios, the lane that minimizes deviation from the 
previous reference trajectory is selected. Terms $\omega_2$--$\omega_4$ 
penalize lateral derivatives to ensure smooth path geometry, while 
$\omega_5$ and $\omega_6$ suppress position and heading discontinuities 
between adjacent discretization steps.

\subsection{Learning-Based Velocity Profile Prediction}\label{subsec3_3:vel_planning}

Longitudinal velocity planning is handled by a learning-based module that provides a traffic-adaptive speed reference for the downstream path optimizer. Rather than predicting full two-dimensional trajectories, the module is deliberately restricted to outputting a sequence of longitudinal velocities. This design choice is motivated by two considerations: first, longitudinal speed behavior in highway driving exhibits stronger statistical regularity than lateral maneuvers, making it more amenable to stable data-driven modeling; second, restricting the learning output to a one-dimensional signal avoids entangling the learning module with safety-critical lateral decisions, which are instead handled with formal guarantees by the MIQP-based planner.

The module adopts VectorNet \cite{gao2020vectornet} as its backbone, retaining its hierarchical graph structure for encoding agent trajectories and map elements as vectorized polylines, and its attention-based global interaction mechanism for capturing multi-vehicle dependencies. The key adaptation lies in the prediction target: rather than predicting future positions, the output layer is reformulated to produce a discretized longitudinal velocity sequence $\mathbf{V}=(v_{1},v_{2},\dots,v_{N})$ over $N$ future time steps. A sigmoid activation is applied at the output layer and scaled to the range $(0,v_{\max})$ based on the road speed limit, ensuring that predicted velocities remain within physically and legally admissible bounds. The corresponding longitudinal position sequence $\mathbf{S}=(s_{1},s_{2},\dots,s_{N})$ is recovered via temporal integration:

\begin{equation}
s_{i}= \sum_{j=0}^{i} v_{j}\Delta t,\quad i=1,2,\dots,N.
\label{eq:s_from_v}
\end{equation}

This position sequence is passed directly to the path planning module as the longitudinal motion reference, maintaining temporal consistency throughout the planning horizon. The model is trained end-to-end using mean squared error loss on real-world highway driving data, with the loss defined as:

\begin{equation}
    L = \frac{1}{N} \sum_{i=1}^{N} (y_i - \hat{y}_i)^2
    \label{eq:pred_loss}
\end{equation}

where $y_i$ and $\hat{y}_i$ denote the ground-truth and predicted longitudinal velocities at time step $i$, respectively.

\section{Experiments}
\label{sec:experiments}

\begin{table*}[!t]  %
\centering
\renewcommand{\arraystretch}{1.2} 
\caption{Parameter configurations}
\label{tab:parameters_values}
\begin{tabular}{@{}ll@{}}
\toprule
Parameters & Values\\
\midrule
$dt$, $\Delta t$, $N$, $M_{\mathrm{tra}}$, $M_{\mathrm{map}}$ & 0.1 s, 0.1 s, 30, 20, 100\\
$\lambda$, $l_{\mathrm{mar}}$, $l_{\mathrm{buf}}$, $\omega_{\mathrm{cell}}$ & 6, 0.3 m, 0.5 m, 0.5\\
$\omega_{1}$, $\omega_{2}$, $\omega_{3}$, $\omega_{4}$, $\omega_{5}$, $\omega_{6}$ & 1, 500, 500, 500, 500, 500\\
$l'_{\min}$, $l''_{\min}$, $l'''_{\min}$ & $-\tan(\pi/3)$, $-3$, $-3$\\
$l'_{\max}$, $l''_{\max}$, $l'''_{\max}$ & $\tan(\pi/3)$, $3$, $3$\\
$W_{EV}$, $L_{EV}$ & 1.8 m, 4.8 m\\
Input and Output Feature Dimensions & 15, 1\\
Learning Rate, Decaying Factor, Batch Size & 0.001, 0.9, 128\\
Loss Function, Optimizer & MSE, Adam\\
Train–Validation–Test Split & 7:2:1\\
$v_{\max}$ & 50 m/s\\
\bottomrule
\end{tabular}
\end{table*}

The H-HTP was implemented in Python, integrating Gurobi v11 \cite{gurobi2024manual} for path optimization and PyTorch for velocity prediction. All experiments were conducted on a computational platform equipped with an Intel i9-9900 CPU (3.60 GHz), 32GB of RAM, and an NVIDIA RTX 2080 Ti GPU, running Ubuntu 22.04. 

Real-world human driving trajectories were sourced from the HighD dataset \cite{krajewski2018highD}, with all test scenarios drawn from a held-out subset that was entirely separate from the training data used for model development.

The experiment adopts a receding-horizon planning strategy, as illustrated in Fig.~\ref{fig:framework}, where the optimal trajectory is re-optimized every $dt$ to reflect the current state. The temporal framework is governed by three key parameters: a unified sampling rate of $0.1~\text{s}$, used both as the trajectory discretization interval ($\Delta t$) and the replanning cycle ($dt$); a $3$-second planning horizon; and a $2$-second observation horizon. These settings correspond to $30$ planning steps ($N = 3~\text{second}/0.1~\text{second}$) and $20$ observation steps ($M_{\mathrm{tra}} = 2~\text{second}/0.1~\text{second}$) for algorithmic implementation. The complete parameter configurations are provided in Table~\ref{tab:parameters_values}.

\subsection{Baseline Methods}
\label{sec:experiments_baseline}
In addition to the H-HTP, we include a learning-only baseline based on the VectorNet architecture~\cite{gao2020vectornet}, referred to as VectorNet-HighD. 
This baseline directly predicts future ego-vehicle trajectories in an end-to-end manner, without downstream optimization or explicit safety constraints.

To ensure a fair comparison, VectorNet-HighD is trained on the same HighD dataset and uses identical input features, observation horizons, and prediction horizons as the learning module in the H-HTP. 
The key difference lies in the output and execution strategy: while the H-HTP predicts a longitudinal velocity profile that serves as a reference for constrained optimization, VectorNet-HighD directly outputs two-dimensional future trajectories.

Section~\ref{sec:experiments_scen} evaluates the trajectory planner in three challenging highway scenarios involving mixed autonomy, where the EV operates autonomously alongside human-driven SVs. Section~\ref{sec:experiments_compt&success} quantitatively analyzes the system’s computational efficiency and solution robustness. Section~\ref{sec:experiments_pred} details the training procedure and empirical performance of the learning-based velocity predictor.

\subsection{Validation on Real-World Scenarios}
\label{sec:experiments_scen}
The HighD dataset involves vehicles traveling at high speeds (25–40 m/s), characteristic of typical highway driving environments. We analyze three representative scenarios in which the EV faces safety challenges. In all cases, the H-HTP successfully generates trajectories that enable the EV to perform emergency evasive maneuvers while maintaining high smoothness and kinematic feasibility. For qualitative comparison, we also report the trajectory predictions of the learning-only baseline VectorNet-HighD under the same scenarios. To ensure a fair evaluation, VectorNet-HighD is executed using the same receding-horizon replanning mechanism as H-HTP, with identical observation windows and replanning intervals. These results validate the H-HTP's effectiveness in safety-critical situations and demonstrate its robustness under extreme conditions. This also suggests its reliability in less complex, lower-risk environments.

\subsubsection{Scenario 1: Frontal cut-in}\label{sec:experiments_scen1}

\begin{figure*}[!t]
  \centering
  \includegraphics[width=.45\linewidth]{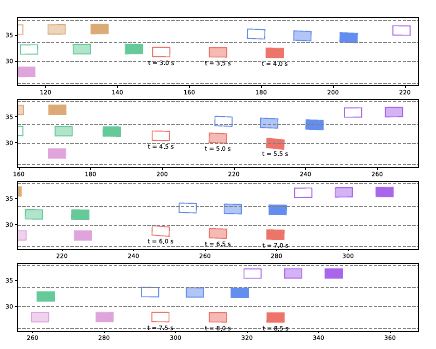}\hfil
  \includegraphics[width=.45\linewidth]{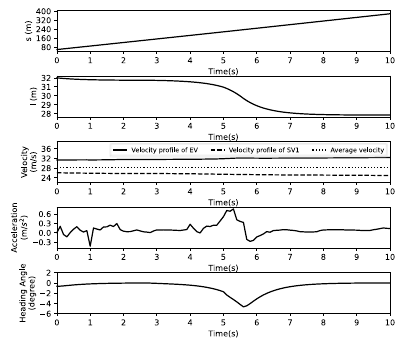}

  \medskip
  \includegraphics[width=.95\linewidth]{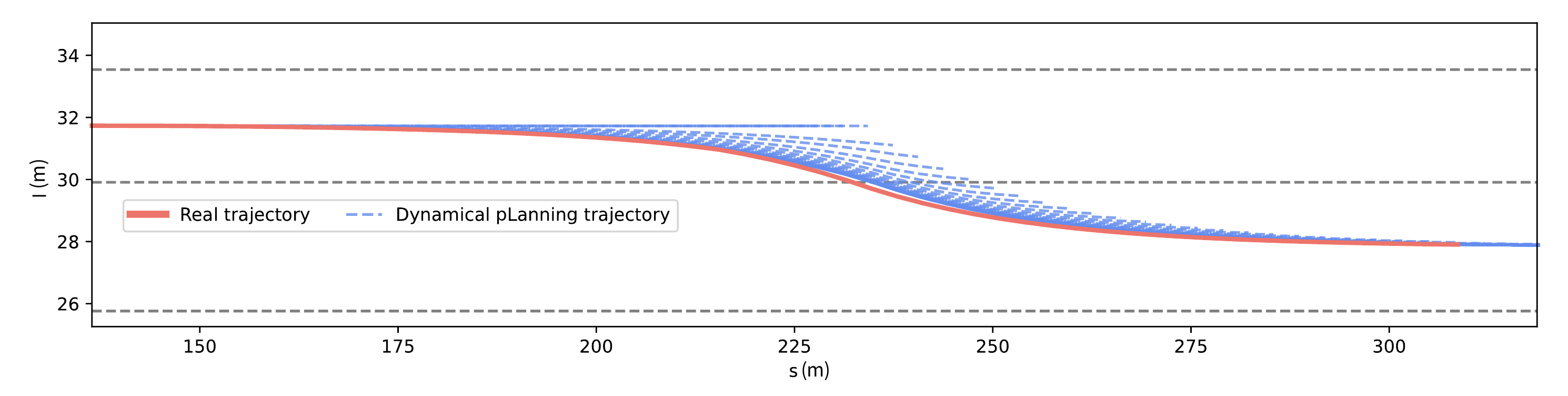}
  
  \caption{Scenario 1 Performance. (a) Sequential snapshots at 0.5-second intervals; (b) kinematic profile showing longitudinal and lateral displacement, velocity, acceleration, and heading angle. In the velocity plot, SV1 corresponds to the blue vehicle shown in (a); (c) the full executed lane-change trajectory along with intermediate trajectory snapshots from online re-planning. (A dynamic visualization is available at \url{https://github.com/alasjia/H-HTP-Framework/tree/main/H_HTP/trajectory_planning/Results_HHTP/gif_displaying_in_paper})}
  \label{fig:scen1}
\end{figure*}

This scenario involves a typical emergency cut-in event, where a slower SV abruptly merges into the EV's lane. Fig.~\ref{fig:scen1} shows the scenario performance with sequential snapshots, kinematic metrics, and lane-change trajectories. The EV is consistently shown in orange-red across all visualizations, while the SVs are depicted using the same set of colors across subsequent figures to ensure clarity.

The SV---shown in blue and located in the left-front position---begins indicating a lane-change intention at $t = \text{3~s}$ and crosses into the EV's lane at $t = \text{4.5~s}$, initiating a collision risk due to the 7~m/s velocity differential. Emergency braking at the EV's current speed of 32~m/s would not only result in high longitudinal jerk and compromised ride comfort, but also significantly increase the risk of rear-end collisions. This risk stems from the combination of short headway and high speed, particularly given the limited reaction time available to drivers of following human-driven vehicles. To mitigate these risks, the H-HTP initiates an early evasive lane change at $t = \text{3~s}$, leveraging 0.1-second replanning cycles to generate dynamically optimized trajectory segments (blue dashed). These segments collectively form a highly smooth, collision-free executed trajectory (orange) that safely avoids the SV within 5~s while preserving passenger comfort---as reflected by the kinematic profile with peak acceleration $\le~\text{0.8~m/s}^2$ and maximum heading angle of $\text{5}^\circ$. This demonstrates the planner's ability to maintain trajectory quality and ride comfort under frequent online updates, even in critical collision scenarios.

\begin{figure*}[!t]
  \centering
  \includegraphics[width=.45\linewidth]{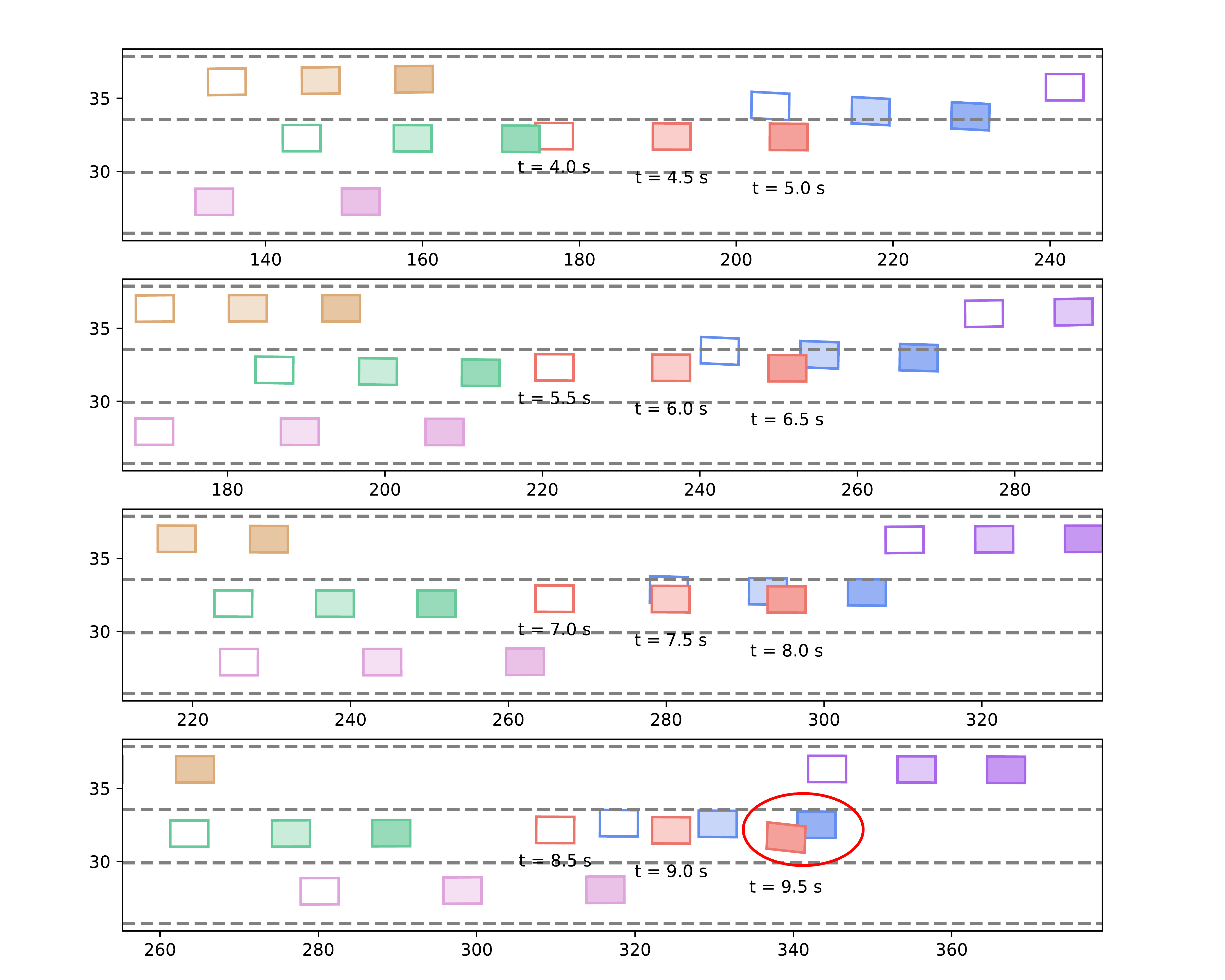}
  \includegraphics[width=.45\linewidth]{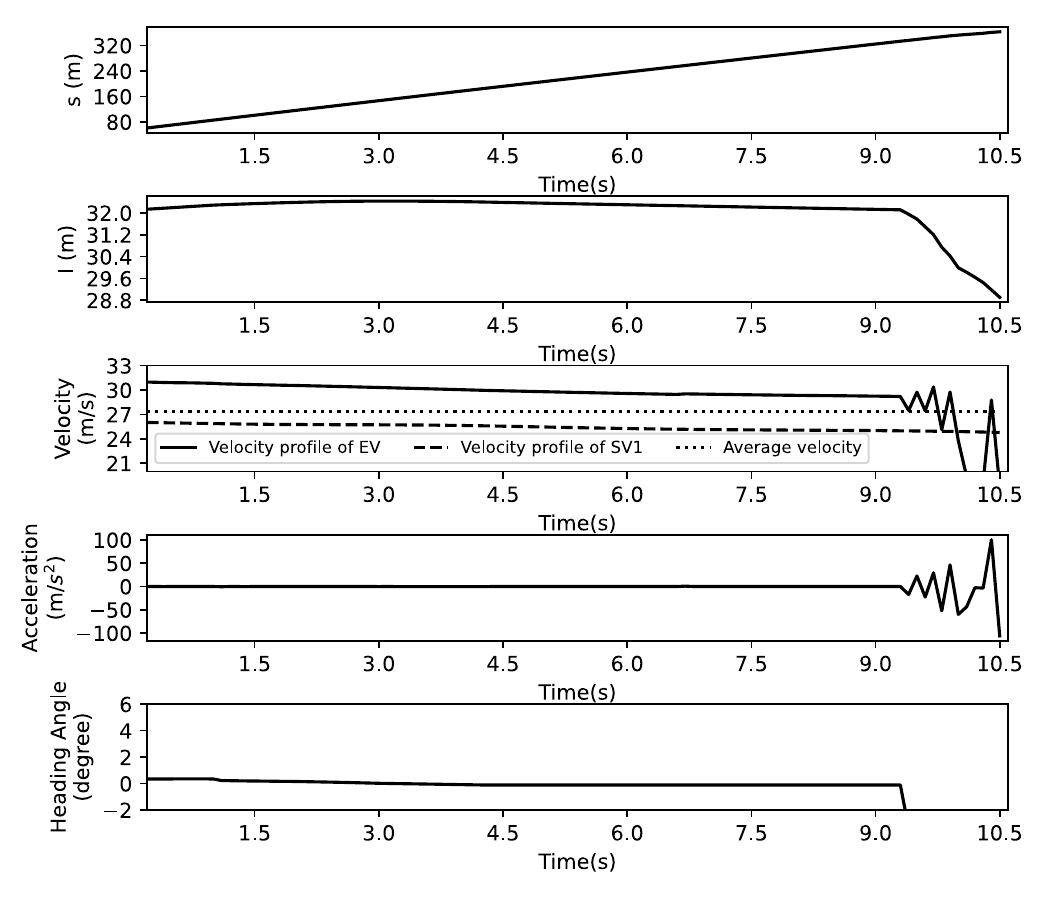}
  
  \caption{Scenario 1 baseline performance (VectorNet-HighD). 
    (a) Predicted motion shown by sequential snapshots at 0.5-second intervals; 
    (b) corresponding kinematic profiles of the predicted trajectory. }

  \label{fig:scen1_e2e}
\end{figure*}

Fig.~\ref{fig:scen1_e2e} further illustrates the qualitative difference between the proposed H-HTP and the VectorNet-HighD under the same observation and prediction horizons. As illustrated by the sequential snapshots, the predicted ego motion exhibits limited lateral adaptation to the rapidly intruding vehicle. The baseline lacks anticipatory awareness of the impending collision risk—it does not initiate a lane-change response until the ego vehicle is on the verge of colliding with the preceding vehicle. By that point, the reaction is too late: the ego vehicle and the intruding vehicle collide between $t = \text{9~s}$ and $t = \text{9.5~s}$ (highlighted with a circled region). The kinematic profiles are consistent with this observation. Once the belated lane-change response is finally triggered, the velocity and acceleration exhibit severe oscillations, with values that far exceed physically reasonable vehicle kinematic limits.

\subsubsection{Scenario 2: Frontal cut-in with close proximity}\label{sec:experiments_scen2}

\begin{figure*}[!t]
  \centering
  \includegraphics[width=.45\linewidth]{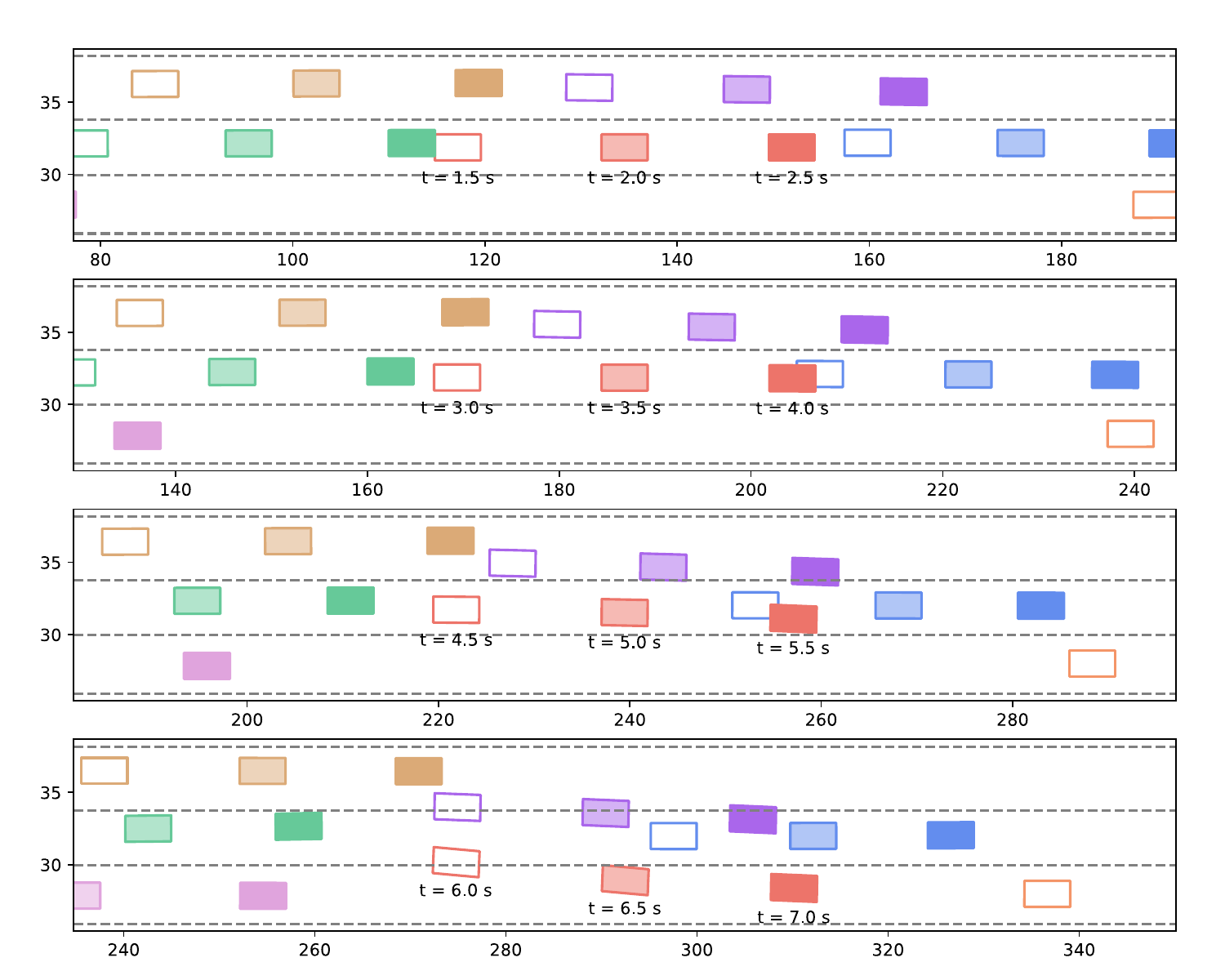}\hfil
  \includegraphics[width=.45\linewidth]{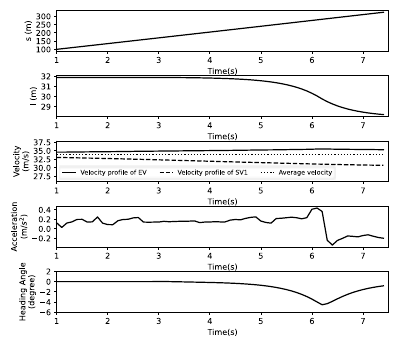}

  \medskip
  \includegraphics[width=.95\linewidth]{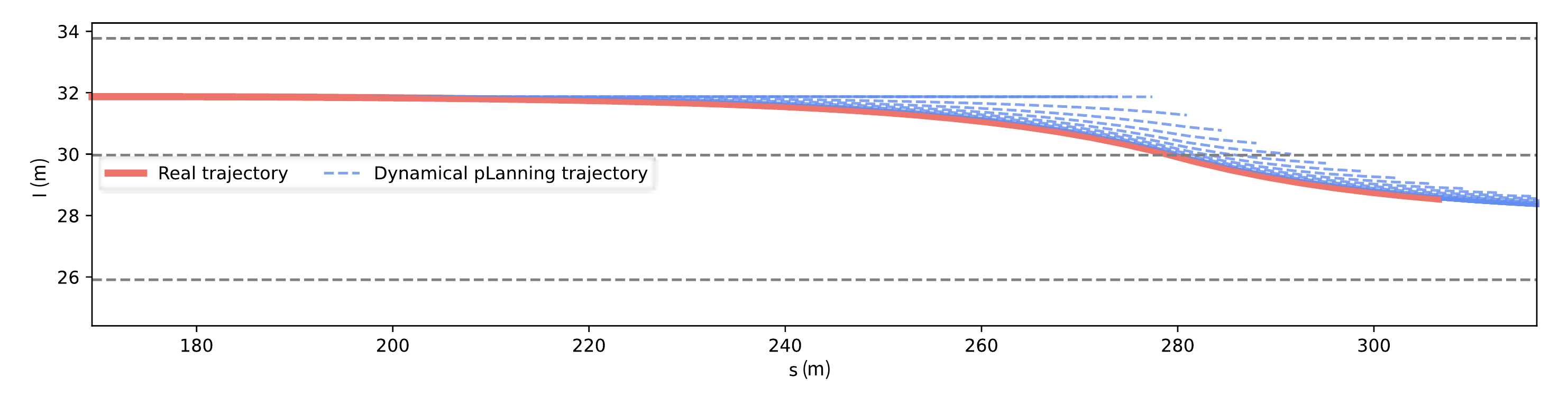}
  
  \caption{Scenario 2 Performance. (a) Sequential snapshots at 0.5-second intervals; (b) kinematic profile showing longitudinal and lateral displacement, velocity, acceleration, and heading angle. In the velocity plot, SV1 corresponds to the purple vehicle shown in (a); (c) the full executed lane-change trajectory along with intermediate trajectory snapshots from online re-planning. (A dynamic visualization is available at \url{https://github.com/alasjia/H-HTP-Framework/tree/main/H_HTP/trajectory_planning/Results_HHTP/gif_displaying_in_paper})}
  \label{fig:scen2}
\end{figure*}

Scenario~2 resembles Scenario~1 but presents a more urgent situation. It involves a slower-moving SV (shown in purple in Fig.~\ref{fig:scen2}~(a)) that is initially positioned in the left-front of the EV and much closer in proximity. At $t = \text{3~s}$, the SV suddenly initiates a hazardous lane change toward the EV’s lane. As it approaches the lane boundary, it becomes nearly parallel to the EV, creating an imminent risk of a sideswipe collision if the EV fails to respond. Meanwhile, emergency braking would also introduce a significant risk of rear-end collision. To mitigate both risks, the H-HTP triggers a defensive lane change to the right at $t = \text{4~s}$, guiding the EV into an adjacent lane while maintaining a nearly constant velocity ($\approx \text{32~m/s}$). The maneuver is completed by $t = \text{7.5~s}$ with sustained safe lateral clearance from the intruding SV. The smooth kinematic response---characterized by a peak acceleration below 0.5~m/s$^2$ and a maximum heading angle of 5$^\circ$, as shown in Fig.~\ref{fig:scen2}~(b) and Fig.~\ref{fig:scen2}~(c)---demonstrates the system’s capability for stable and comfortable lateral control in time-critical interactions.

\begin{figure*}[!t]
  \centering
  \includegraphics[width=.45\linewidth]{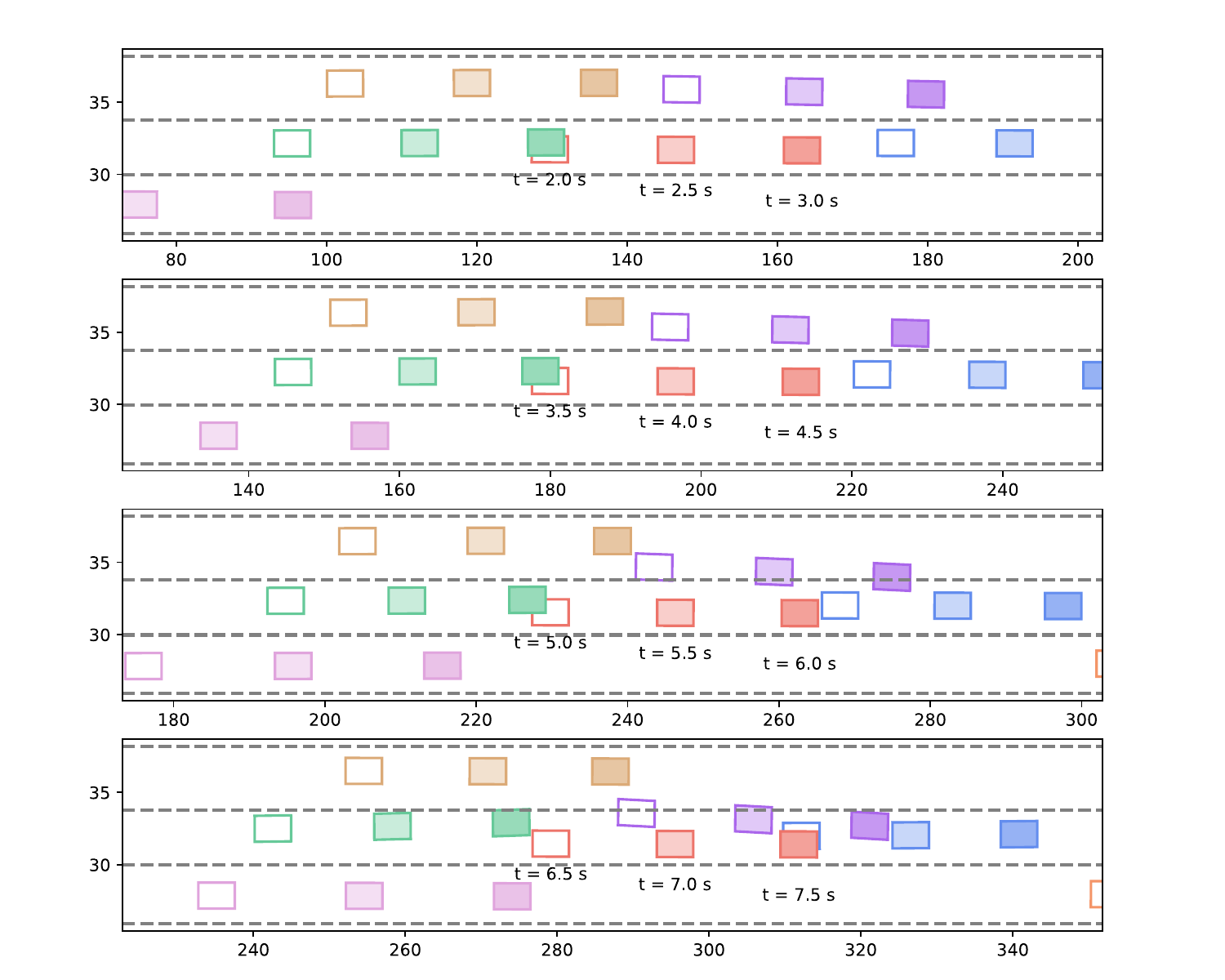}
  \includegraphics[width=.45\linewidth]{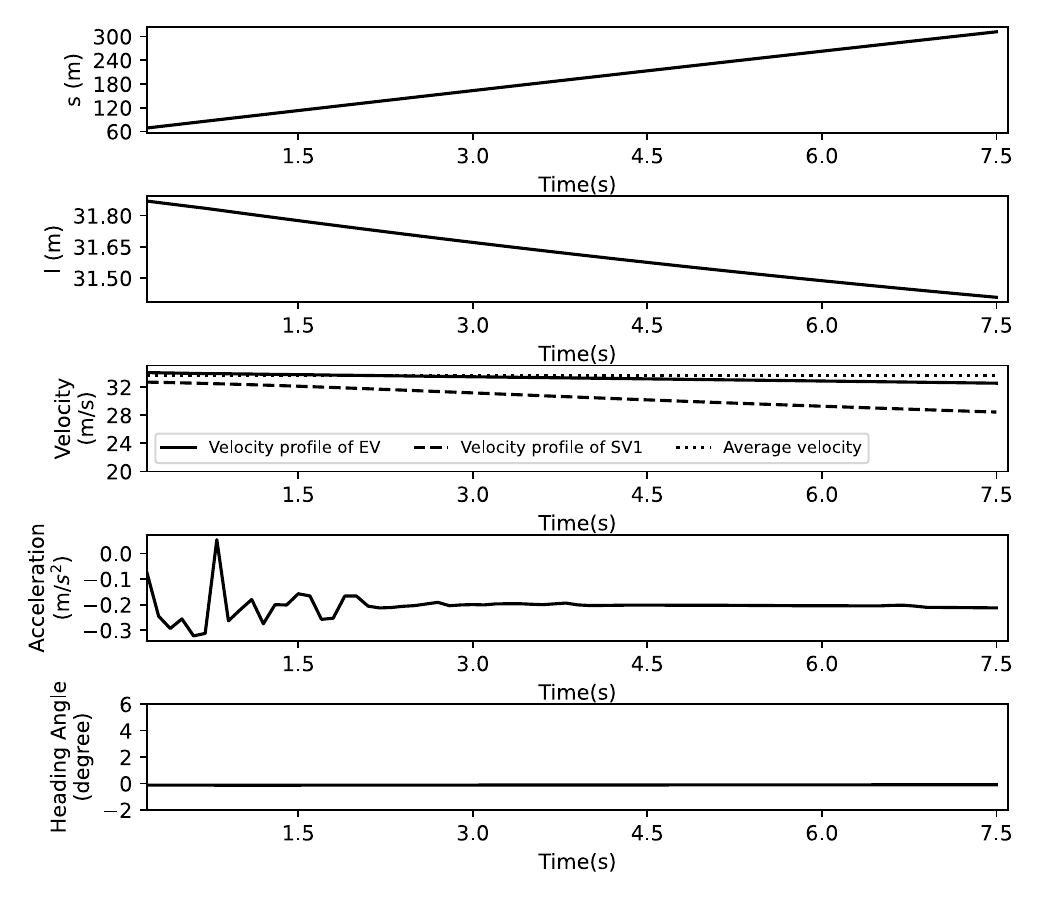}
  
  \caption{Scenario 2 baseline performance (VectorNet-HighD). 
    (a) Predicted motion shown by sequential snapshots at 0.5-second intervals; 
    (b) corresponding kinematic profiles of the predicted trajectory. }
  \label{fig:scen2_e2e}
\end{figure*}

Fig.~\ref{fig:scen2_e2e} reports the VectorNet-HighD baseline results for Scenario~2. The baseline prediction in this case shows no behavioral adaptation to the hazardous interaction. As illustrated in the sequential snapshots, the predicted ego trajectory remains lane-keeping throughout the horizon, despite the cut-in vehicle rapidly intruding. Although no collision occurs within the time range provided by the dataset, the final moments reveal a high-risk state in which the ego vehicle is rapidly approaching the preceding vehicle with an already critically small separation distance. The kinematic profiles are consistent with this observation. The lateral displacement remains nearly monotonic and the heading angle shows little variation.

\subsubsection{Scenario 3: Rear cut-in}\label{sec:experiments_scen3}

\begin{figure*}[!t]
  \centering
  \includegraphics[width=.45\linewidth]{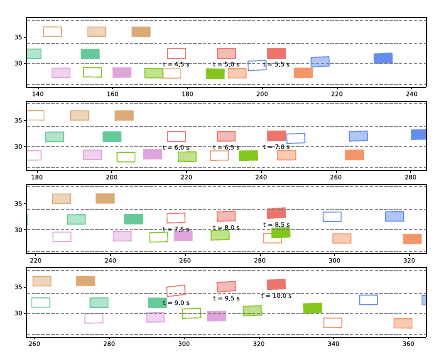}\hfil
  \includegraphics[width=.45\linewidth]{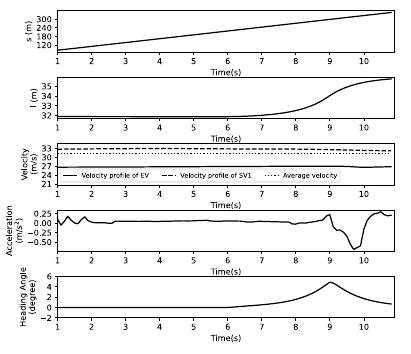}

  \medskip
  \includegraphics[width=.95\linewidth]{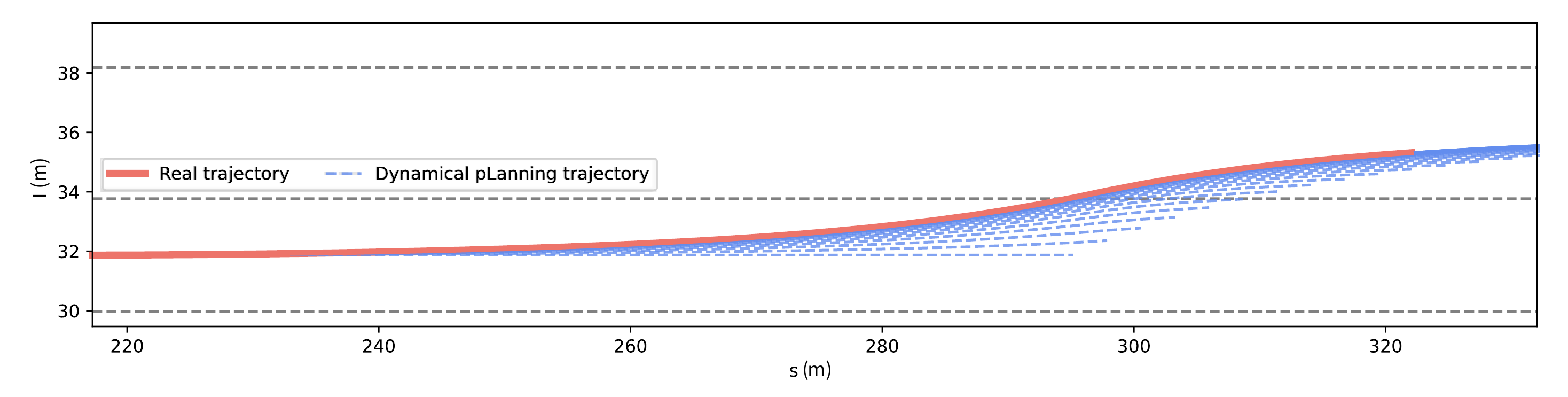}
  
  \caption{Scenario 3 Performance. (a) Sequential snapshots at 0.5-second intervals; (b) kinematic profile showing longitudinal and lateral displacement, velocity, acceleration, and heading angle. In the velocity plot, SV1 corresponds to the green vehicle shown in (a); (c) the full executed lane-change trajectory along with intermediate trajectory snapshots from online re-planning. (A dynamic visualization is available at \url{https://github.com/alasjia/H-HTP-Framework/tree/main/H_HTP/trajectory_planning/Results_HHTP/gif_displaying_in_paper})}
  \label{fig:scen3}
\end{figure*}

Scenario~3 addresses a dynamic cut-in event involving a SV (shown in green in Fig.~\ref{fig:scen3}~(a) approaching from the EV’s right-rear at 33~m/s. At $t = \text{6~s}$, the SV suddenly initiates a lane change into the EV’s lane, creating an immediate sideswipe-collision risk. In response, the planner promptly initiates an evasive left-lane change into an adjacent lane with sufficient forward clearance, completing the avoidance maneuver within 4~s. As shown in Figs.~\ref{fig:scen3}~(b) -- Figs.~\ref{fig:scen3}~(c), the EV maintains smooth and stable motion throughout, with all kinematic metrics remaining within safe operational limits. This result further confirms the planner’s robustness in handling fast-developing, high-risk scenarios via real-time trajectory adaptation.

\begin{figure*}[!t]
  \centering
  \includegraphics[width=.45\linewidth]{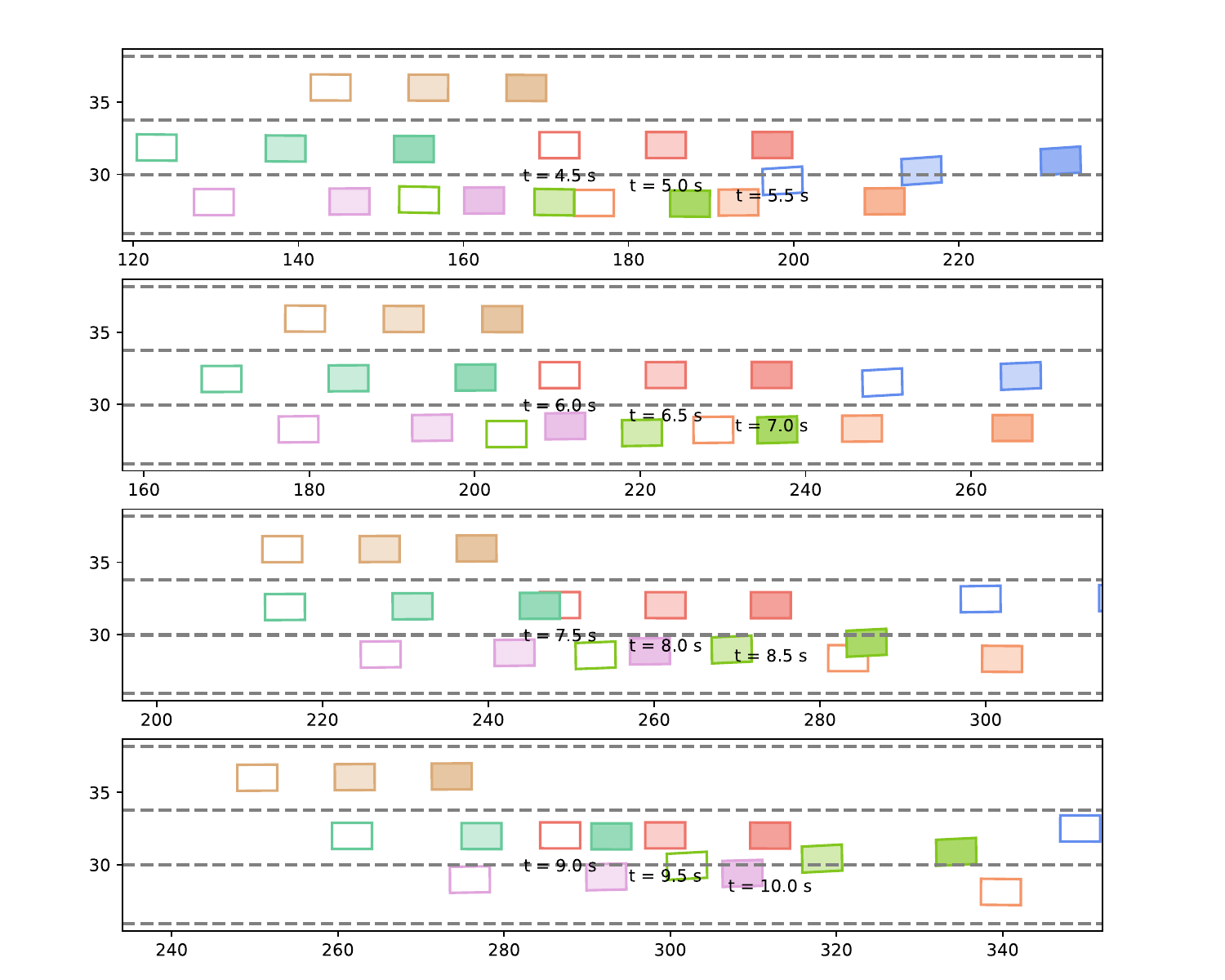}
  \includegraphics[width=.45\linewidth]{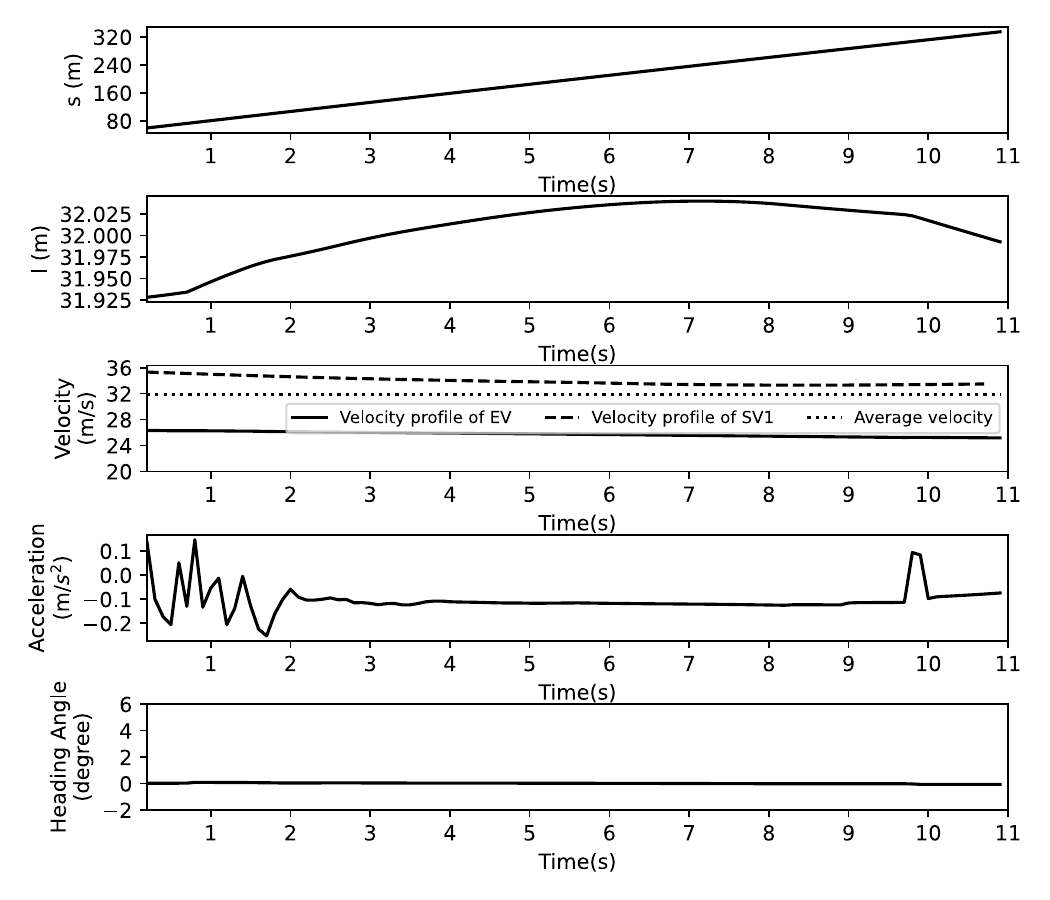}
  
  \caption{Scenario 3 baseline performance (VectorNet-HighD). 
    (a) Predicted motion shown by sequential snapshots at 0.5-second intervals; 
    (b) corresponding kinematic profiles of the predicted trajectory. }
  \label{fig:scen3_e2e}
\end{figure*}

Fig.~\ref{fig:scen3_e2e} presents the VectorNet-HighD baseline results for Scenario~3. In this scenario, the baseline performs comparatively better, maintaining lane-keeping behavior while sustaining an adequate distance from the intruding vehicle throughout the interaction. This is attributable to the baseline adopting a lower velocity compared to the H-HTP. While this conservative speed avoids a collision, it comes at the cost of reduced driving efficiency. Similar to Scenario~2, the kinematic profiles reflect a normal lane-keeping process; however, minor anomalous oscillations in the acceleration are observed during the initial time steps. This may indicate inherent instability in the full-trajectory prediction approach.

The above observations suggest that the VectorNet-HighD baseline often fails to reproduce the lateral evasive behaviors required for effective obstacle avoidance. This observation highlights the intrinsic difficulty of learning lateral decisions purely from data, particularly in rare and time-critical interactions. The proposed H-HTP framework enables proactive, safety-oriented maneuvers through constrained optimization and online replanning, which is crucial for handling rare but time-critical frontal cut-in events.

\subsection{Computational Efficiency and Success Rate}
\label{sec:experiments_compt&success}

The proposed H-HTP was validated through 591,213 receding-horizon planning cycles across 6,900 driving scenarios extracted from recordings 53–55 of the HighD dataset. 

\begin{figure}[!t]
  \centering
  \includegraphics[width=.48\linewidth]{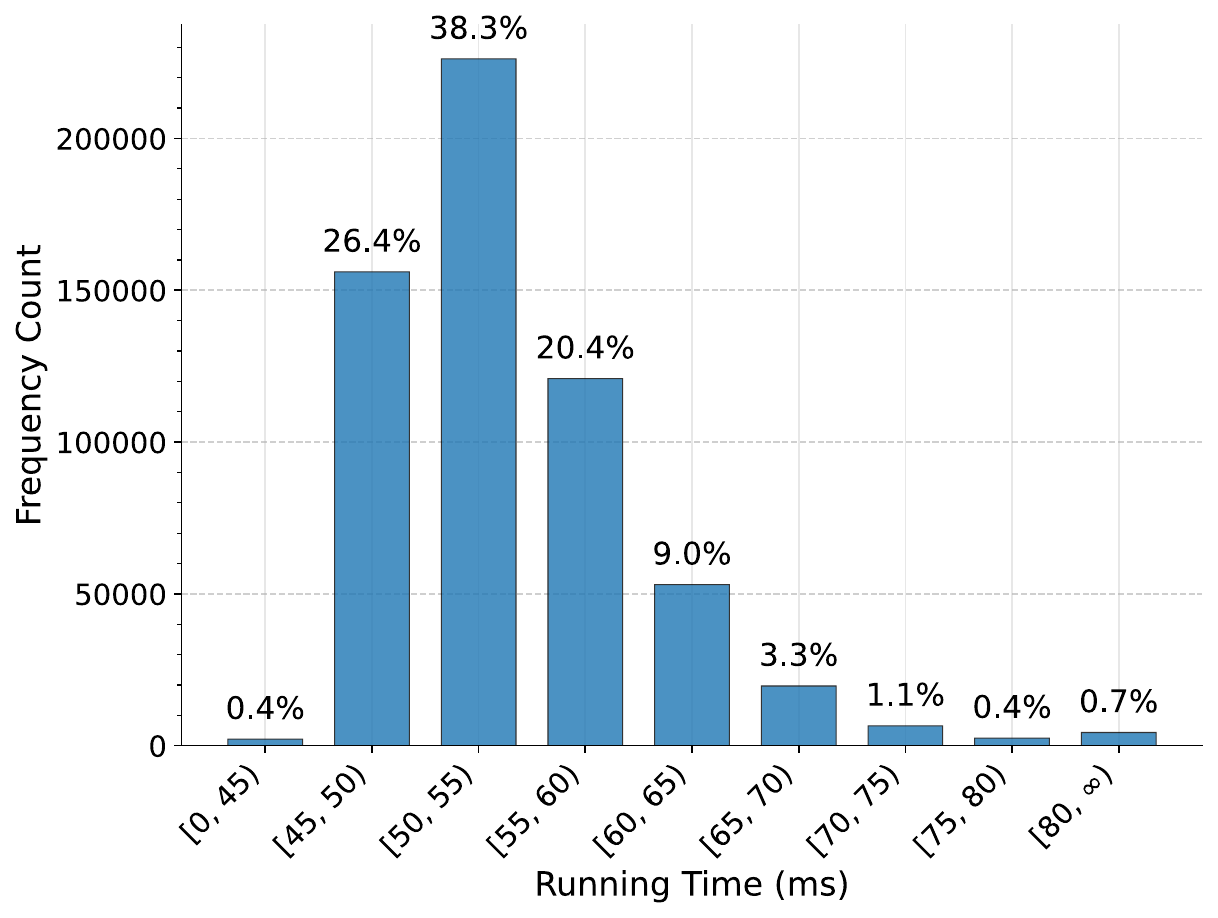}\hfil
  \includegraphics[width=.48\linewidth]{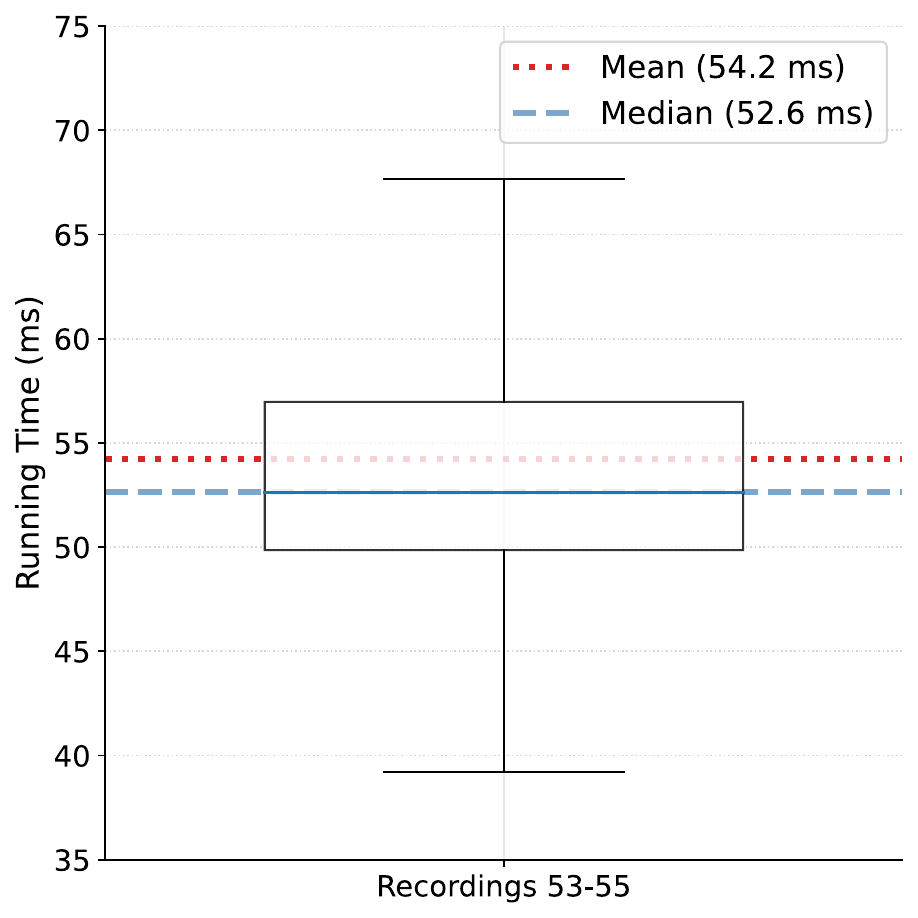}
  \caption{Planning cycle runtime statistics.
  The left panel shows the distribution of planning cycle runtimes, while the right panel presents the corresponding box plot.
  For clarity, extremely small ($<45$~ms) and large ($>80$~ms) values are grouped at the axis limits, and outliers are omitted.}
  \label{fig:cycle_rt}
\end{figure}

\begin{figure}[!t]
  \centering
  \includegraphics[width=.48\linewidth]{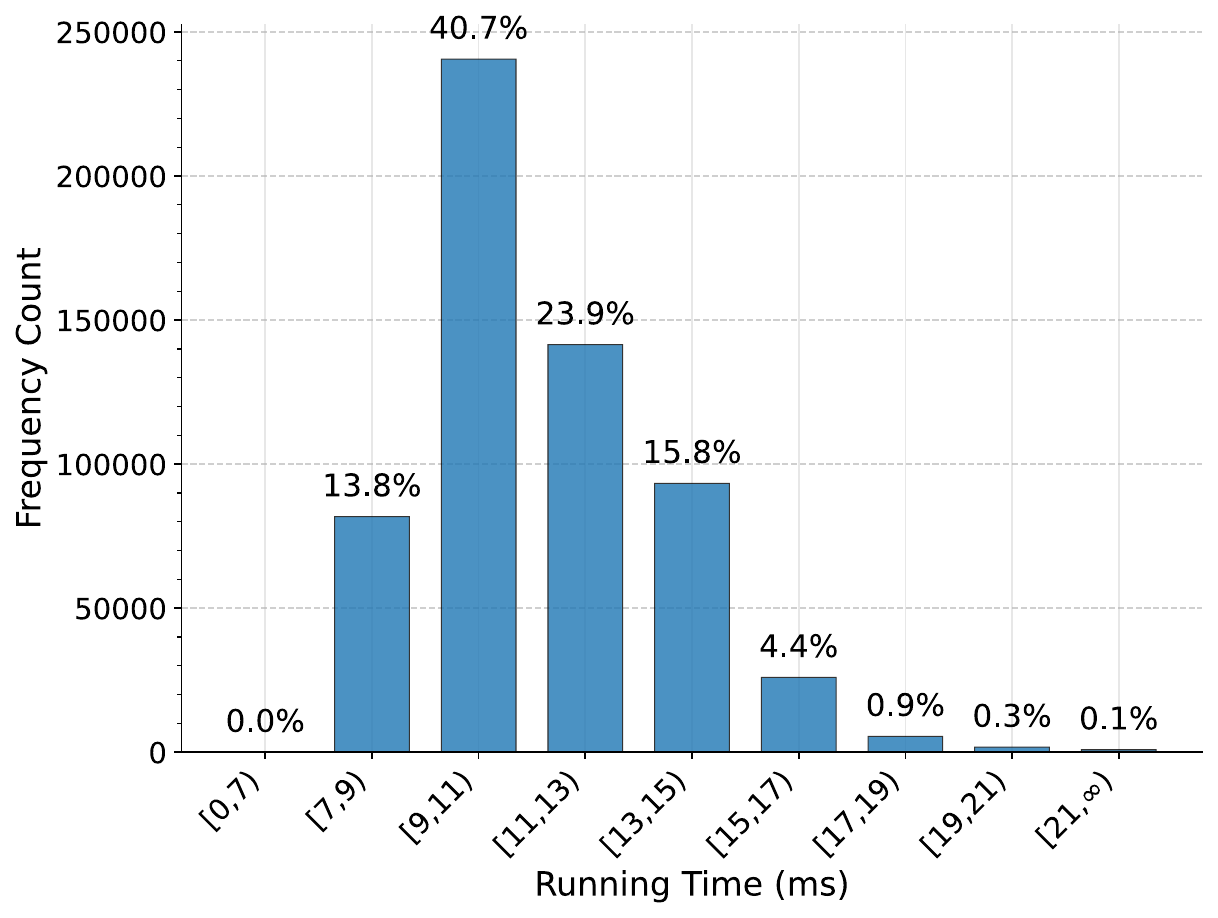}\hfil
  \includegraphics[width=.48\linewidth]{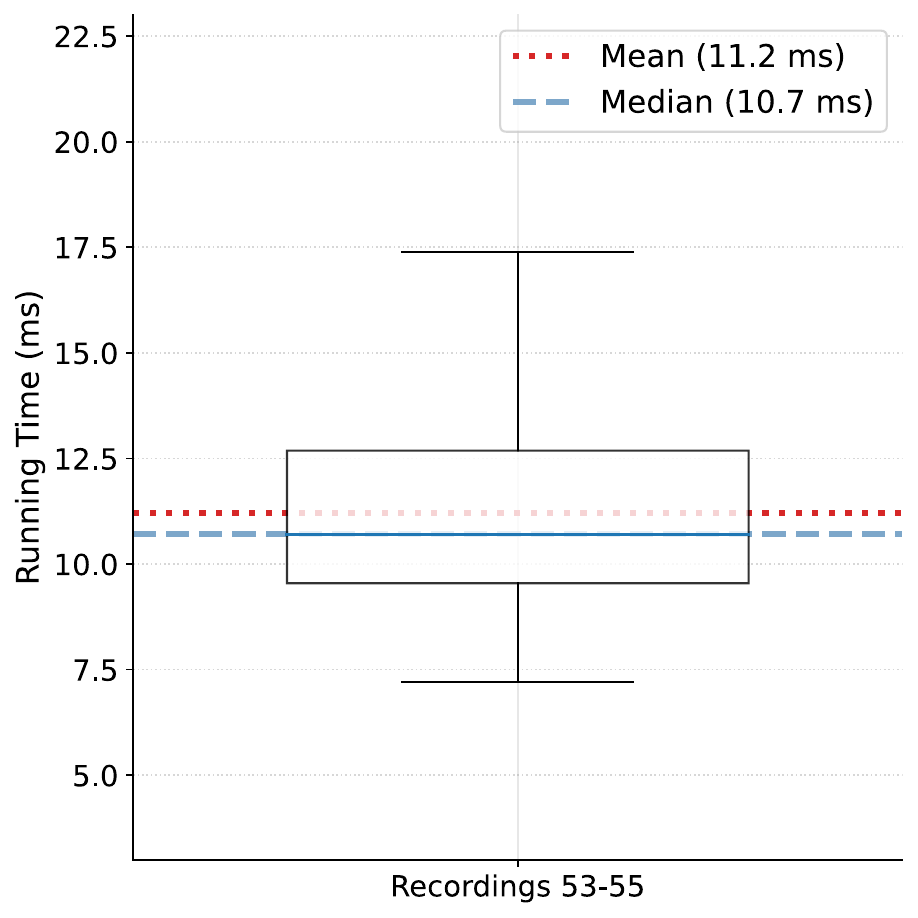}
  \caption{Path optimization runtime statistics.
  The left panel illustrates the distribution of optimization runtimes, and the right panel shows the corresponding box plot.
  For visualization clarity, values exceeding $21$~ms are capped at the axis limit and outliers are omitted.}
  \label{fig:opt_rt}
\end{figure}

We report runtime statistics measured on the same hardware platform described in Section \ref{sec:experiments}, excluding visualization and disk I/O. The planning-cycle runtime refers to the wall-clock time required to generate one executed trajectory segment, including velocity prediction, MIQP-based path optimization, and state updates. As shown in Fig.~\ref{fig:cycle_rt}, the runtime is mainly distributed within 45–65 ms (mean: 54.2 ms; median: 52.6 ms). Within each cycle, the MIQP solver time (Gurobi) accounts for 7–15 ms (mean: 11.2 ms; median: 10.7 ms), as shown in Fig.~\ref{fig:opt_rt}. These results indicate that the proposed hybrid pipeline can operate in real time under a 0.1 s replanning period.

To put the computational cost of learning-only inference into perspective, we additionally report the forward-pass runtime of the VectorNet-HighD baseline under the same observation horizon and input features. On the same platform, VectorNet-HighD requires 20.4 ms per inference (mean, batch size = 1), which is lower than the full planning-cycle time of H-HTP but does not account for any downstream feasibility enforcement or collision-avoidance constraints.

Two success criteria were used to evaluate the robustness of H-HTP. The scenario success rate measures the ability to continuously generate feasible trajectories over the full ground-truth horizon of a scenario, while the cycle success rate reflects feasibility at a single replanning cycle. As reported in Table~\ref{tab:success_rates}, H-HTP achieves a 97.0\% scenario success rate with consistently high cycle-level feasibility across recordings 53–55.

These results indicate that the proposed hybrid framework can reliably maintain feasibility under conservative MIQP formulations with strict safety margins, outperforming representative optimization-based planning approaches reported in the literature.

\begin{table}
\centering
\caption{SUCCESS RATES ACROSS RECORDINGS}
\label{tab:success_rates}

\begin{tabular}{cccc}
\toprule
Recording &  & Scenario & Cycle \\
Index     & Scenarios & Success & Success \\
          &           & Rate (\%) & Rate (\%) \\
\midrule
53 & 2421 & 96.1 & 98.1 \\
54 & 2313 & 97.5 & 98.7 \\
55 & 2166 & 97.3 & 98.8 \\
\bottomrule
\end{tabular}

\end{table}

\subsection{Training and Evaluation of the Velocity Prediction Module}
\label{sec:experiments_pred}

\begin{figure}
    \centering
    \includegraphics[width=\linewidth]{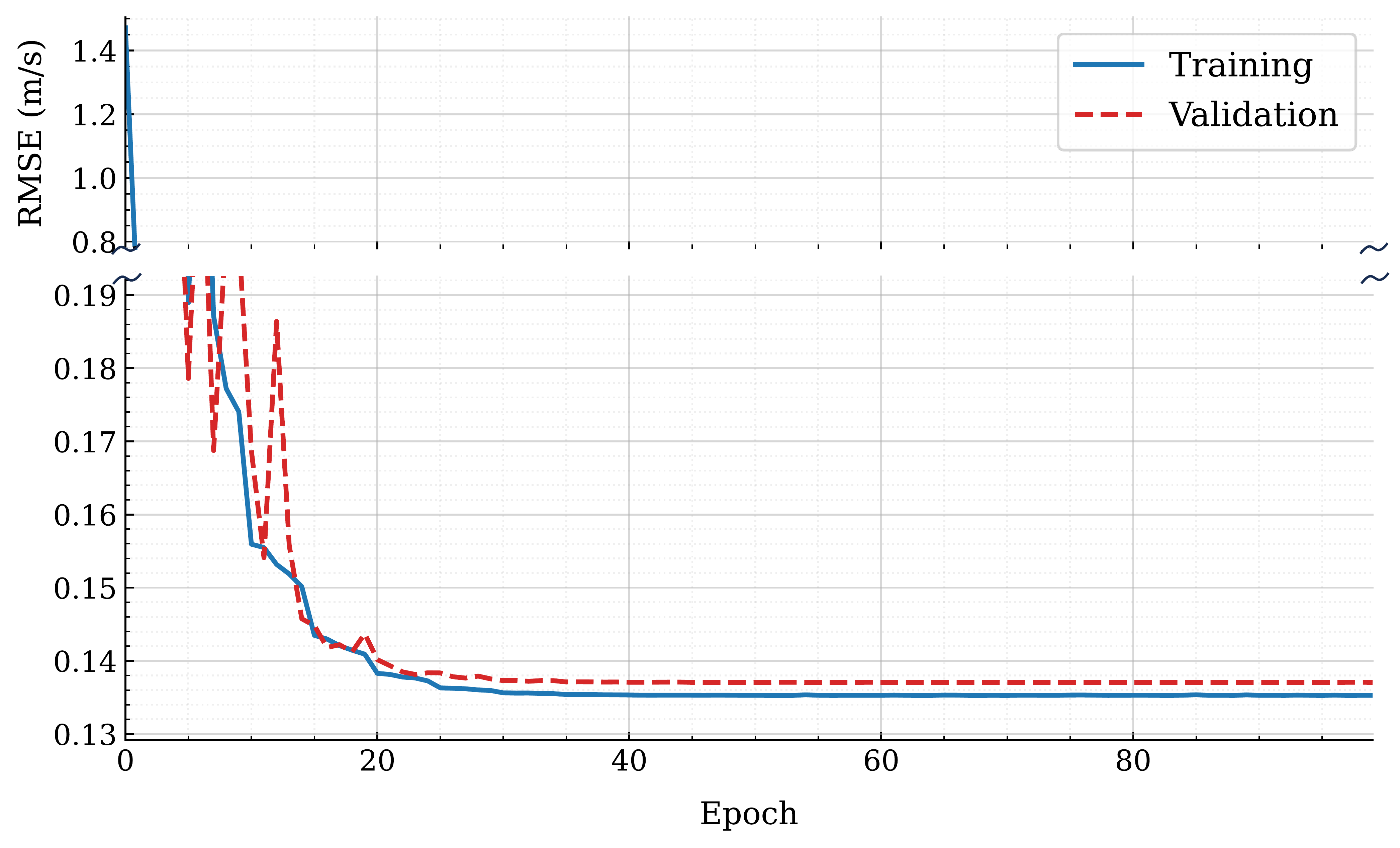}
    \caption{Training Convergence of the Velocity Prediction Module (H-HTP)}
    \label{fig:Prediction_Error}
\end{figure}

\begin{figure}
    \centering
    \includegraphics[width=\linewidth]{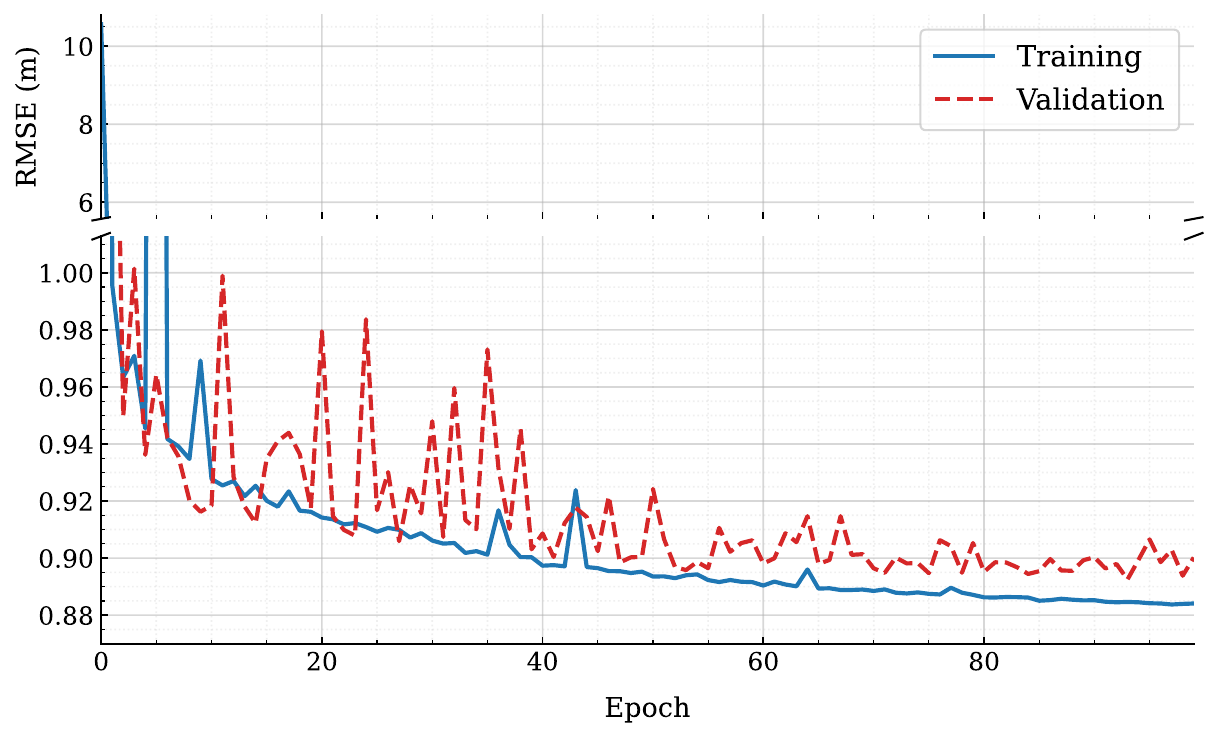}
    \caption{Training Convergence of the Velocity Prediction Module (VectorNet-HighD)}
    \label{fig:Prediction_Error_e2e}
\end{figure}

The neural network architecture was trained using the hyperparameters listed in Table~\ref{tab:parameters_values}. Trajectory data were extracted from three-lane HighD scenarios (recordings 26–45) and split into training (317,588 samples), validation (90,739 samples), and test (45,371 samples) sets in a 7:2:1 ratio. As shown in Fig.~\ref{fig:Prediction_Error} and Fig.~\ref{fig:Prediction_Error_e2e}, both the proposed velocity prediction module and the VectorNet-HighD baseline exhibit stable convergence behavior during training, with the proposed model achieving smoother and more consistent loss profiles.

The model began to stabilize after approximately 20 epochs. To address kinematic irregularities observed during preliminary validation, training was extended to 100 epochs. The additional iterations improved motion smoothness without compromising prediction accuracy, resulting in smoother acceleration profiles that are particularly beneficial for downstream comfort-aware trajectory planning.

After training, the model was evaluated on recordings 54–56, which comprise 56,550 trajectories extracted using a 1-second sliding window. Since the proposed model outputs discrete longitudinal velocity sequences rather than full trajectories, predicted velocity profiles are transformed into positional coordinates via temporal integration using equation \eqref{eq:s_from_v}, enabling the computation of displacement error (DE). As presented in Table~\ref{tab:DE_comparison}, the proposed velocity planning module yields substantially lower trajectory-level displacement errors than VectorNet-HighD under the same data distribution. This improvement reflects the benefit of focusing the learning task on longitudinal kinematics, where the VectorNet backbone can be more effectively leveraged, while delegating lateral feasibility to the optimization-based planner.

\begin{table}
\centering
\caption{Trajectory-Level Displacement Errors for Planning-Oriented Motion Prediction}
\label{tab:DE_comparison}

\begin{tabular}{@{}lcc@{}}
\toprule
 & \makecell{VectorNet-HighD} & \makecell{Velocity Prediction\\(H-HTP)} \\
\midrule
Epochs & 100 & 100 \\
Prediction Target & Trajectory & Velocity profile \\
ADE (m) & 0.66 & 0.06 \\
FDE (m) & 1.52 & 0.16 \\
\bottomrule
\end{tabular}

\end{table}

\section{Conclusions}
\label{sec:conclusions}

This paper presents H-HTP, a hybrid trajectory planning framework for autonomous highway driving that combines learning-based adaptability with optimization-based formal safety guarantees. The framework introduces a principled division of labor: a learning module provides a traffic-adaptive velocity reference, while a constrained MIQP-based path planner ensures collision avoidance and kinematic feasibility. A linearization strategy reduces the number of integer variables to one per time step, enabling real-time execution without sacrificing hard safety constraints. Experiments on the HighD dataset across 6,900 scenarios and over 591,000 planning cycles demonstrate a scenario success rate above 97\% with an average planning-cycle time of approximately 54 ms. Compared with the learning-only baseline, H-HTP exhibits substantially stronger lateral adaptability in safety-critical cut-in scenarios, reliably producing smooth, kinematically feasible, and collision-free trajectories.

The current framework assumes accurate surrounding vehicle predictions and is evaluated only on structured highway geometry. Future work will address prediction uncertainty in safety corridor construction and extend the framework to more diverse road topologies.

\bibliography{refs}

\end{document}